%% file: main.tex
\documentclass[final]{cvpr}

\usepackage{times}
\usepackage{epsfig}
\usepackage{graphicx}
\usepackage{amsmath}
\usepackage{amssymb}

\usepackage[font=small,skip=6pt]{caption}
\usepackage[labelformat=simple,font={small}]{subcaption}
\usepackage[flushleft]{threeparttable}

\usepackage[dvipsnames]{xcolor}
\definecolor{citecolor}{HTML}{0071bc}
\usepackage[pagebackref=true,breaklinks=true,colorlinks,citecolor=citecolor,bookmarks=false]{hyperref}
\usepackage[author={Max Schlepzig}]{pdfcomment}
\usepackage{color, colortbl}
\usepackage{bm}
\usepackage{booktabs}
\usepackage{multirow}

\usepackage{float}
\usepackage{stfloats}
\def\cvprPaperID{3431} 
\def\confYear{CVPR 2021}

\definecolor{ForestGreen}{rgb}{0.13, 0.55, 0.13}
\definecolor{ForestGreen2}{rgb}{0.2, 0.5372549019607843, 0.1803921568627451}
\definecolor{newGREEN}{RGB}{34,139,34}
\definecolor{myMaroon}{RGB}{231, 52, 52}
\definecolor{Maroon}{rgb}{0.69, 0.19, 0.0}
\definecolor{my_cyan}{RGB}{112, 242, 244}

\newcommand{\FG}[1]{\textcolor{ForestGreen}{\textbf{#1}}}

\newcommand{\Gd}{\rowcolor{gray!45}}

\newcommand{\veryshortarrow}[1][3pt]{\mathrel{%
 \hbox{\rule[\dimexpr\fontdimen22\textfont2-.2pt\relax]{#1}{.4pt}}%
 \mkern-4mu\hbox{\usefont{U}{lasy}{m}{n}\symbol{41}}}}

\newcommand{\dd}{\mathop{}\!\mathrm{d}}
\newenvironment{packed_itemize}{
	\begin{itemize}
		\setlength{\itemsep}{0pt}
		\setlength{\parskip}{0pt}
		\setlength{\parsep}{0pt}
	}{\end{itemize}}

\begin{document}
\title{Prototypical Cross-domain Self-supervised Learning for Few-shot \\ Unsupervised Domain Adaptation}


\author{Xiangyu Yue$^{1,}$\thanks{Equal contribution; correspondence to \texttt{xyyue@berkeley.edu}}\quad Zangwei Zheng$^{2,}$\footnotemark[1]\quad Shanghang Zhang$^{1}$\quad Yang Gao$^{3}$ \\Trevor Darrell$^{1}$\quad Kurt Keutzer$^{1}$\quad Alberto Sangiovanni Vincentelli$^{1}$
\vspace{1mm}
\\
$^1$UC Berkeley\quad $^2$Nanjing University\quad $^3$Tsinghua University 
}
\maketitle

\begin{abstract}
Unsupervised Domain Adaptation (UDA) transfers predictive models from a fully-labeled source domain to an unlabeled target domain. In some applications, however, it is expensive even to collect labels in the source domain, making most previous works impractical. To cope with this problem, recent work performed instance-wise cross-domain self-supervised learning, 
followed by an additional fine-tuning stage. 
However, the instance-wise self-supervised learning only learns and aligns low-level discriminative features. 
In this paper, we propose an end-to-end Prototypical Cross-domain Self-Supervised Learning (PCS) framework for Few-shot Unsupervised Domain Adaptation (FUDA)\footnote{Project page: \url{http://xyue.io/pcs-fuda}}. PCS not only performs cross-domain low-level feature alignment, but it also encodes and aligns semantic structures in the shared embedding space across domains. Our framework captures category-wise semantic structures of the data by in-domain prototypical contrastive learning; and performs feature alignment through cross-domain prototypical self-supervision. Compared with state-of-the-art methods, PCS improves the mean classification accuracy over different domain pairs on FUDA by 10.5\%, 3.5\%, 9.0\%, and 13.2\% on Office, Office-Home, VisDA-2017, and DomainNet, respectively. 
\end{abstract}

\section{Introduction}
Deep Learning has achieved remarkable performance in various computer vision tasks, such as image classification~\cite{he2016deep, huang2017densely} and semantic segmentation~\cite{long2015fully, zhao2017pyramid, chen2017deeplab}. Despite high accuracy, deep neural networks trained on specific datasets often fail to generalize to new domains owing to the \textit{domain shift} problem~\cite{torralba2011unbiased,donahue2014decaf,tzeng2017adversarial}. Unsupervised domain adaptation (UDA) transfers  predictive models from a fully-labeled source domain to an unlabeled target domain. Although it is challenging with no label information in the target domain, many UDA methods~\cite{tzeng2017adversarial, hoffman2018cycada, long2015learning, ganin2016domain} could achieve high accuracy on the target domain using the abundant explicit supervision in source domain, together with the unlabeled target samples for domain alignment. 

\begin{figure}[t]
 \centering
 \includegraphics[width=3.2in, trim={0cm 0cm 0cm 0cm}]{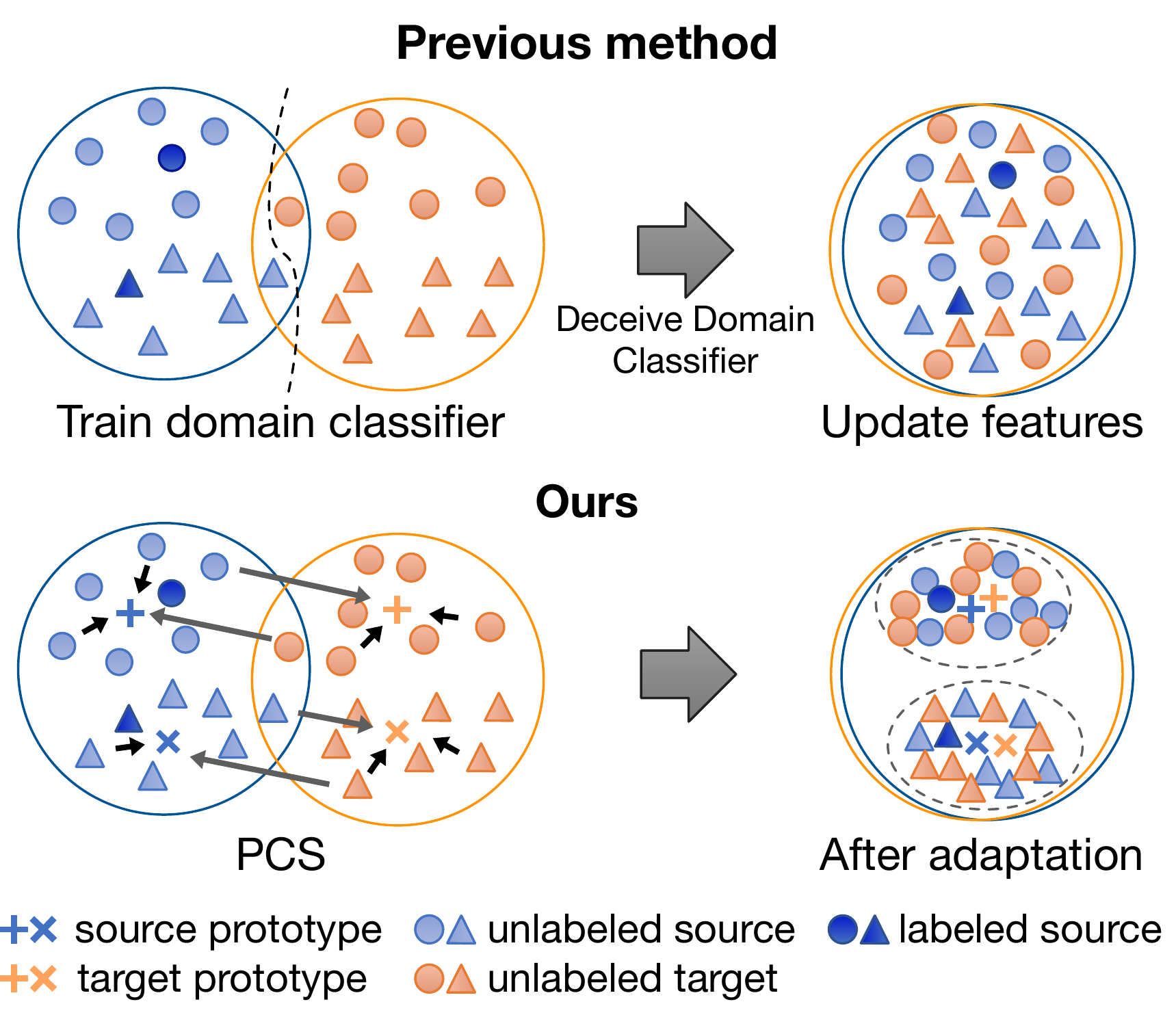}
 \caption{We address the task of few-shot unsupervised domain adaptation. Top: Existing domain-classifier based methods align source and target distributions but fail to extract discriminative features due to lack of labeled data. Bottom: Our method estimates prototypes for in-domain and cross-domain 
 self-supervised learning to extract domain-aligned discriminative features. }
 \label{fig:teaser}
\end{figure}

In some real-world applications, however, providing large-scale annotations even in the source domain is often challenging due to the high cost and difficulty of annotation. Taking medical imaging for instance, each image of the Diabetic Retinopathy dataset~\cite{gulshan2016development} is annotated by a panel of 7 or 8 U.S. board-certified ophthalmologists, with a total group of 54 doctors. Thus practically it is too stringent to assume the availability of source data with abundant labels.

In this paper, to cope with the labeling costs of the source domain, we instead consider a few-shot unsupervised domain adaptation (FUDA) setting, where only an extremely small fraction of source samples are labeled, while all the rest source and target samples remain unlabeled. Most state-of-the-art UDA methods align source and target features by minimizing some form of distribution distances~\cite{long2015learning, long2018conditional, sun2017correlation, ganin2016domain}, and learn the discriminative representation by minimizing the supervision loss on fully-labeled source domain data. In FUDA, however, since we have a very limited number of labeled source samples, it is much harder to learn discriminative features in the source domain, not to mention in the target domain. 

Several recent papers~\cite{he2020momentum,chen2020simple,grill2020bootstrap, oord2018representation, wu2018unsupervised} on self-supervised learning (SSL) present promising representation learning results on images from a single domain and \cite{kim2020cross} further extended to perform SSL across two domains for better domain adaptation performance. Despite the improved performance, the instance-based method in~\cite{kim2020cross} has some fundamental weaknesses. First, the semantic structure of the data is not encoded by the learned structure. This is because the in-domain self-supervision in \cite{kim2020cross} treats two instances as negative pairs as long as they are from different samples, regardless of the semantic similarity. Consequently, many instances sharing the same semantic are undesirably pushed apart in the feature space. Second, the cross-domain instance-to-instance matching in~\cite{kim2020cross} is very sensitive to abnormal samples. Imagine a case where the embeddings of source and target samples are far apart (\textit{i.e.} the domain gap is large) and one abnormal source sample is mapped closer to all target samples than any other source sample. Then the method in~\cite{kim2020cross} would match all target samples to the same source sample (cf. Figure~\ref{fig:id_vs_proto}). For a given sample, the matched sample in the other domain may change drastically during the training procedure, making the optimization harder to converge. Third, the two-stage pipeline (\textit{i.e.} SSL followed by domain adaptation) is complicated and experiments show that the optimal DA methods for different datasets are different. As a result, the training is rather cumbersome and it is unclear how to choose the optimal DA method in the second stage for different datasets.

In this paper, we propose \textit{Prototypical Cross-domain Self-supervised learning}, a novel single-stage framework for FUDA that unifies representation learning and domain alignment with few-shot labeled source samples. PCS contains three major components to learn both discriminative and domain-invariant features. First, PCS performs in-domain prototypical self-supervision to implicitly encode the semantic structure of data into the embedding space. This is motivated by~\cite{li2020prototypical}, but we further leverage the known semantic information of the task and learn better semantic structure in each domain. 
Second, PCS performs cross-domain instance-to-prototype matching to transfer knowledge from source to the target in a more robust manner. Instead of instance-to-instance matching, matching a sample to a prototype (\textit{i.e. }representative embedding for a group of instances that are semantically similar) is more robust to abnormal instances in the other domain and makes the optimization converge faster and more smoothly. 
Third, PCS unifies prototype learning with cosine classifier and update cosine classifier adaptively with source and target prototypes. 
transfers from source prototypes to target prototypes for better performance on the target domain. 
In order to further mitigate the effect of cross-domain mismatching, we perform entropy maximization to obtain a more diversified output. We show that together with entropy minimization, this is equivalent to maximizing the mutual information (MI) between input image and the network prediction. 

To summarize, our contributions are three-fold:
\vspace{-1.3mm}
\begin{packed_itemize}
 \item We propose a novel Prototypical Cross-domain Self-supervised learning framework (PCS) for few-shot unsupervised Domain Adaptation. 
 \item We propose to leverage prototypes to perform better semantic structure learning, discriminative feature learning, and cross-domain alignment in a unified, unsupervised and adaptive manner. 
 \item While it is hard to choose the optimal domain adaptation method in the complex two-stage framework~\cite{kim2020cross}, PCS can be easily trained in an end-to-end matter, and outperforms all state-of-the-art methods by a large margin across multiple benchmark datasets. 
\end{packed_itemize}




\section{Related Work}
\noindent\textbf{Domain Adaptation.}
Unsupervised Domain Adaptation (UDA)~\cite{gopalan2011domain} addresses the problem of transferring knowledge from a fully-labeled source domain to an unlabeled target domain. Most UDA methods have focused on feature distribution alignment. Discrepancy-based methods explicitly compute the Maximum Mean Discrepancy (MMD)~\cite{gretton2012kernel} between the source and the target to align the two domains~\cite{long2015learning, tzeng2014deep, long2016unsupervised}. Long \etal~\cite{long2017deep} proposed to align the joint distributions using the Joint MMD criterion. Sun \etal~\cite{sun2017correlation} and Zhuo \etal~\cite{zhuo2017deep} further proposed to align second-order statistics of source and target features. With the development of Generative Adversarial Networks~\cite{goodfellow2014generative}, additional papers proposed to perform domain alignment in the feature space with adversarial learning~\cite{ganin2015unsupervised, tzeng2017adversarial, hoffman2018cycada, xie2018learning, long2018conditional, shen2017wasserstein}. Recently, image translation methods, \textit{e.g.}~\cite{zhu2017unpaired,liu2016coupled} have been adopted to further improve domain adaptation by performing pixel-level alignment~\cite{hoffman2018cycada, bousmalis2017unsupervised, russo2018source, murez2018image, yue2019domain, sankaranarayanan2018generate, shrivastava2017learning}. Instead of explicit feature alignment, Saito \etal~\cite{saito2019semi} proposed minimax entropy for adaptation. 
While these methods have full supervision on the source domain, similar to~\cite{kim2020cross}, we focus on a new adaptation setting with few source labels. 

\begin{figure*}[t]
 \centering
 \includegraphics[width=6.85in, trim={0.0cm 0cm 0cm 0cm}]{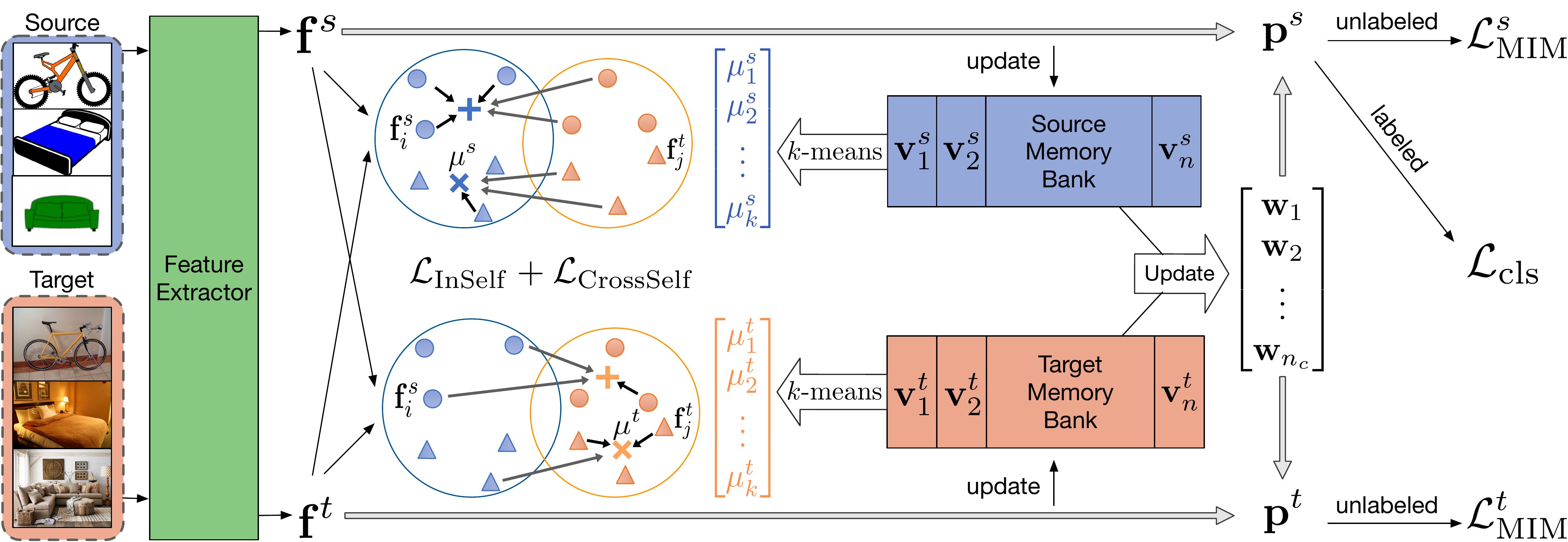}
 \caption{An overview of the PCS framework. In-domain and cross-domain self-supervision are performed between normalized feature vectors $\mathbf{f}$ and prototypes $\mathbf{\mu}$ computed by clustering vectors $\mathbf{v}$ in memory banks. Features with confident predictions ($\mathbf{p}$) are used to adaptively update classifier vectors $\mathbf{w}$. MI maximization and classification loss are further used to extract discriminative features. }
 \label{fig:framework}
\end{figure*} 

\noindent\textbf{Self-supervised Learning.} Self-supervised learning (SSL) is a subset of unsupervised learning methods where supervision is automatically generated from the data~\cite{jing2020self, doersch2015unsupervised, zhang2016colorful,noroozi2016unsupervised,gidaris2018unsupervised,wang2020pre}. One of the most common strategies for SSL is handcrafting auxiliary pretext tasks predicting future, missing or contextual information. In particular, image colorization~\cite{zhang2016colorful, larsson2016learning}, patch location prediction~\cite{doersch2015unsupervised, doersch2017multi}, image jigsaw puzzle~\cite{noroozi2016unsupervised}, image inpainting~\cite{pathak2016context} and geometric transformations~\cite{gidaris2018unsupervised,dosovitskiy2014discriminative} have been shown to be helpful. 
Currently, contrastive learning~\cite{bachman2019learning, he2020momentum, oord2018representation,tian2019contrastive, misra2020self} has achieved state-of-the-art performance on representation learning~\cite{grill2020bootstrap, chen2020simple, chen2020improved, chen2020big}. Most contrastive methods are instance-wise, aiming to learn an embedding space where samples from the same instance are pulled closer and samples from different instances are pushed apart~\cite{wu2018unsupervised, chen2020simple}. Recently, contrastive learning based on prototypes has shown promising results in representation learning~\cite{li2020prototypical, asano2020self, caron2020unsupervised, garnot2020metric}.

\noindent\textbf{Self-supervised Learning for Domain Adaptation.}
Self-supervision-based methods incorporate SSL losses into the original task network~\cite{ghifary2015domain, ghifary2016deep}. Reconstruction was first utilized as self-supervised task in some early works~\cite{ghifary2015domain, ghifary2016deep}, in which source and target share the same encoder to extract domain-invariant features. 
To capture both domain-specific and shared properties, Bousmalis \etal~\cite{bousmalis2017unsupervised} explicitly extracts image representations into two spaces, one private for each domain and one shared across domains.
In~\cite{carlucci2019domain}, solving jigsaw puzzle~\cite{noroozi2016unsupervised} was leveraged as a self-supervision task to solve domain adaptation and generalization. Sun \etal~\cite{sun2019unsupervised} further proposed to perform adaptation by jointly learning multiple self-supervision tasks. The feature encoder is shared by both source and target images, and the extracted features are then fed into different self-supervision task heads. 
Recently, based on instance discrimination~\cite{wu2018unsupervised}, Kim~\etal~\cite{kim2020cross} proposed a cross-domain SSL approach for adaptation with few source labels. SSL has also been incorporated for adaptation in other fields, including point cloud recognition~\cite{achituve2020self}, medical imaging~\cite{ihler2020self}, action segmentation~\cite{chen2020action}, robotics~\cite{jeong2020self}, facial tracking~\cite{yoon2019self}, \textit{etc}.

\section{Approach}

In few-shot unsupervised domain adaptation, we are given a very limited number of labeled source images $\mathcal{D}_{s} = \left\{\left(\mathbf{x}_i^s, y_i^s\right)\right\}_{i=1}^{N_s}$, as well as unlabeled source images $\mathcal{D}_{su} = \left\{\left(\mathbf{x}_i^{su}\right)\right\}_{i=1}^{N_{su}}$. In the target domain, we are only given unlabeled target images $\mathcal{D}_{tu} = \left\{\left(\mathbf{x}_i^{tu}\right)\right\}_{i=1}^{N_{tu}}$. The goal is to train a model on $\mathcal{D}_s, \mathcal{D}_{su}, \text{and } \mathcal{D}_{tu}$; and evaluate on $\mathcal{D}_{tu}$.

The base model consists of a feature encoder $F$, a $\ell_2$ normalization layer, which outputs a normalized feature vector $\mathbf{f} \in \mathbb{R}^d$ and a cosine similarity-based classifier $C$. 

\subsection{In-domain Prototypical Contrastive Learning}

We learn a shared feature encoder $F$ that extracts discriminative features in both domains. Instance Discrimination~\cite{wu2018unsupervised} is employed in~\cite{kim2020cross} to learn discriminative features. As an instance-wise contrastive learning method, it results in an embedding space where all instances are well separated. Despite promising results, instance discrimination has a fundamental weakness: the semantic structure of the data is not encoded by the learned representations. This is because two instances are treated as negative pairs as long as they are from different samples, regardless of their semantics. 
For a single domain, ProtoNCE~\cite{li2020prototypical} is proposed to learn semantic structure of the data by performing iterative clustering and representation learning. The goal is to drive features within the same cluster to become more aggregated and features in different clusters further apart. 

However, naively applying ProtoNCE to $\mathcal{D}_s\cup\mathcal{D}_{su} \cup\mathcal{D}_{tu}$ in our domain adaptation setting would cause potential problems. Primarily due to the domain shift, images of different classes from different domains can be incorrectly aggregated into the same cluster, and images of the same class from different domains can be mapped into clusters that are far apart. To mitigate these problems, we propose to perform prototypical contrastive learning separately in $\mathcal{D}_s\cup\mathcal{D}_{su}$ and in $\mathcal{D}_{tu}$. This aims to prevent the incorrect clustering of images across domains and indiscriminative feature learning. 

Specifically, two memory banks \bm{$V^s$} and \bm{$V^t$} are maintained for source and target respectively:
\begin{equation}
\bm{V^s} = [\mathbf{v}_1^s, \cdots, \mathbf{v}_{(N_s+N_{su})}^s], \,\, \bm{V^t} = [\mathbf{v}_1^t, \cdots, \mathbf{v}_{N_{tu}}^t],
\end{equation}
where $\mathbf{v}_i$ is the stored feature vector of $\mathbf{x}_i$, initialized with $\mathbf{f}_i$ and updated with a momentum $m$ after each batch:
\begin{equation}
\mathbf{v}_i \leftarrow m\mathbf{v}_i + (1-m) \mathbf{f}_i.
\end{equation}
In order for in-domain prototypical contrastive learning, $k$-means clustering is performed on \bm{$V^s$} and \bm{$V^t$} to get source clusters $\bm{C}^s = \{C_1^{(s)}, C_2^{(s)}, \dots, C_k^{(s)}\}$ and similarly $\bm{C}^t$
with normalized source prototypes $\{\mu_j^s\}_{j=1}^k$ and normalized target prototypes $\{\mu_j^t\}_{j=1}^k$. 
Specifically, $\mu_j^s = \frac{\mathbf{u}^s_j}{\|\mathbf{u}^s_j\|}$, where $\mathbf{u}_j^s = \frac{1}{|C^{(s)}_{j}|}\sum_{\mathbf{v}^s_i \in C^{(s)}_{j}} \mathbf{v}^s_i$. 
We explain only on the source domain for succinct notation, all operations are performed on target similarly.

During training, with the feature encoder $F$, we compute a feature vector $\mathbf{f}^s_i = F(\mathbf{x}_i^s)$. To perform in-domain prototypical contrastive learning, we compute the similarity distribution vector between $\mathbf{f}_i^s$ and $\{\mu_j^s\}_{j=1}^k$ as $P^s_{i} = [P^s_{i,1},P^s_{i,2},\dots, P^s_{i,k}] $, with 

\begin{equation}
 P^s_{i,j} = \frac{\exp (\mu_j^s \cdot \mathbf{f}_i^s / \phi) }{\sum_{r=1}^k \exp(\mu_r^s \cdot \mathbf{f}_i^s / \phi)},
\end{equation}
where $\phi$ is a temperature value determining the level of concentration. Then the in-domain prototypical contrastive loss can be written as:
\begin{equation}
 \mathcal{L}_\mathrm{PC} =\sum_{i=1}^{N_s + N_{su}} \mathcal{L}_{CE} (P_i^s, c_s(i)) + \sum_{i=1}^{N_{tu}} \mathcal{L}_{CE} (P_i^t, c_t(i)) 
\end{equation}
where $c_s(\cdot)$ and $c_t(\cdot)$ return the cluster index of the instance. 

Due to the randomness in clustering, we perform $k$-means on the samples $M$ times with different number of clusters $\{k_m\}_{m=1}^M$. Moreover, in the FUDA setting, since the number of classes $n_c$ is known, we set $k_m = n_c$ for most $m$.
The overall loss for in-domain self-supervision is:
\begin{equation}
 \mathcal{L}_\mathrm{InSelf} = \frac{1}{M} \sum_{m=1}^M \mathcal{L}_\mathrm{PC}^{(m)}
\end{equation}

\begin{figure}[t]
 \centering
 \includegraphics[width=3.2in, trim={0cm 0cm 0cm 0cm}]{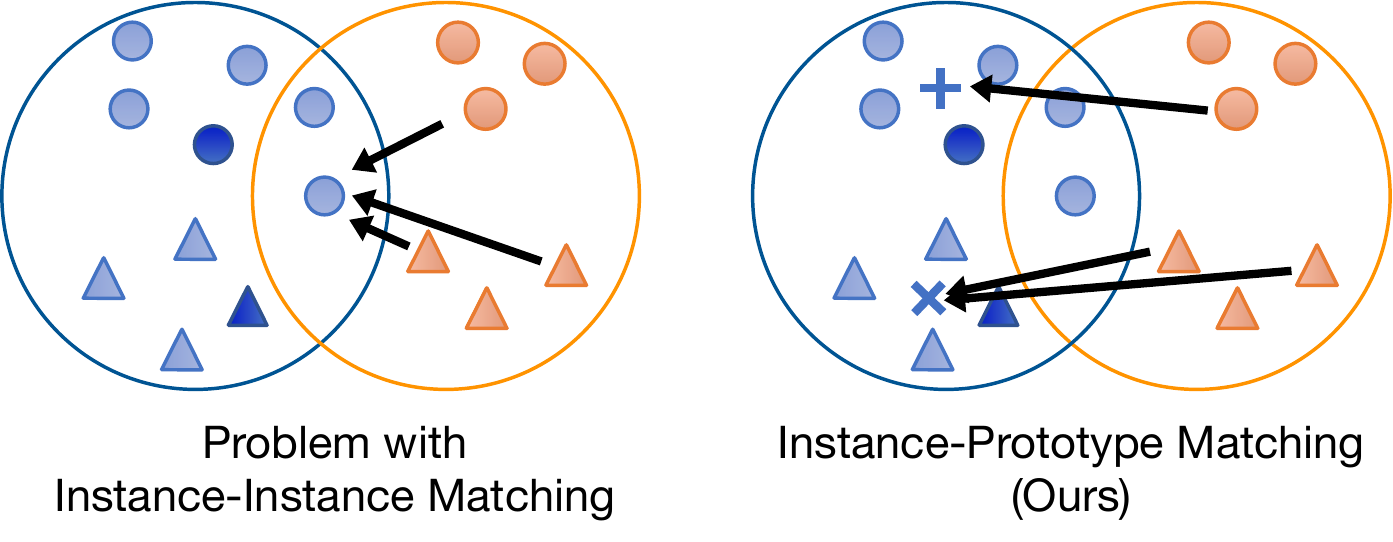}
 \caption{Comparison of cross-domain instance-instance (I-I) matching~\cite{kim2020cross} (left) and our cross-domain instance-prototype (I-P) matching (right). Left: I-I incorrectly matches all orange samples to the same blue sample. Right: I-P robustly matches samples to the correct prototypes. }
 \label{fig:id_vs_proto}
\end{figure}

\subsection{Cross-domain Instance-Prototype SSL}

In order to explicitly enforce learning domain-aligned and more discriminative features in both source and target domains, we perform cross-domain instance-prototype self-supervised learning. 

Many previous works focus on domain alignment via discrepancy minimization or adversarial learning. However, these methods provide inferior performance or have unstable training. Moreover, most of them focus on distribution matching, without considering semantic similarity matching across domains. Instance-instance matching~\cite{kim2020cross} is proposed to match an instance $i$ to another instance $j$ in the other domain with the most similar representation. However, due to the domain gap, instances can be easily mapped to instances of different classes in the other domain. In some cases, if an outlier in one domain is extremely close to the other domain, it will be matched to all the instances in the other domain, as illustrated in Figure~\ref{fig:id_vs_proto}. 

Instead, our method discovers positive matching as well as negative matchings between instance and cluster prototypes in different domains. To find a matching for an instance $i$, we perform entropy minimization on the similarity distribution vector between its representation, \textit{e.g.} $\mathbf{f}_i^s$ and the centroids of the other domain, \textit{e.g.} $\{\mu_j^t\}_{j=1}^k$. 

Specifically, given feature vector $\mathbf{f}_i^s$ in the source domain, and centroids $\{\mu_j^t\}_{j=1}^k$ in the target domain, we first compute the similarity distribution vector $P_{i}^{s\veryshortarrow t} = [P_{i,1}^{s\veryshortarrow t}, \dots, P_{i,k}^{s\veryshortarrow t}]$, in which 
\begin{equation}
 P_{i,j}^{s\veryshortarrow t} = \frac{\exp(\mu_j^t \cdot \mathbf{f}_i^s / \tau)}
 {\sum_{r=1}^k\exp( \mu_r^t \cdot \mathbf{f}_i^s / \tau )}.
\end{equation}
Then we minimize the entropy of $P_{i}^{s\veryshortarrow t}$, which is:
\begin{equation}
 H(P_{i}^{s\veryshortarrow t}) = -\sum_{j=1}^k P_{i,j}^{s\veryshortarrow t}\log P_{i,j}^{s\veryshortarrow t}.
\end{equation}
Similarly, we can compute $H(P_{i}^{t\veryshortarrow s})$, and the final loss for cross-domain instance-prototype SSL is: 
\begin{equation}
 \mathcal{L}_\mathrm{CrossSelf} = \sum_{i=1}^{N_s+N_{su}} H(P_{i}^{s\veryshortarrow t}) + \sum_{i=1}^{N_{tu}} H(P_{i}^{t\veryshortarrow s})
\end{equation}

\subsection{Adaptive Prototypical Classifier Learning}
The goal of this section is to learn a better domain-aligned, discriminative feature encoder $F$ and more importantly, a cosine classifier $C$ that could achieve high accuracy on the target domain. 

The cosine classifier $C$ consists of weight vectors $\mathbf{W} = [\mathbf{w}_1, \mathbf{w}_2, \dots, \mathbf{w}_{n_c}]$, where $n_c$ denotes the total number of classes, and a temperature $T$. The output of $C$, $\small{\frac{1}{T}\mathbf{W^{\mathrm{T}}}\mathbf{f}}$ is fed into a softmax layer $\sigma$ to obtain the final probabilistic output $\mathbf{p(x)} = \sigma(\frac{1}{T} \mathbf{W^{\mathrm{T}}}\mathbf{f})$. With the availability of the labeled set $\mathcal{D}_{s}$, it is straightforward to train $F$ and $C$ with a standard cross-entropy loss for classification:
\begin{equation}
 \mathcal{L}_\mathrm{cls} = \mathbb{E}_{(\mathbf{x}, y)\in \mathcal{D}_s} \mathcal{L}_{CE}(\mathbf{p(x)}, y)
\end{equation}
However, since $\mathcal{D}_{s}$ is quite small under FUDA setting, only training with $\mathcal{L}_\mathrm{cls}$ is hard to get a $C$ with high performance on the target. 

\paragraph{Adaptive Prototype-Classifier Update (APCU)} Note that for $C$ to classify samples correctly, the direction of a weight vector $\mathbf{w}_i$ needs to be representative of features of the corresponding class $i$. This indicates that the meaning of $\mathbf{w}_i$ coincide with the ideal cluster prototype of class $i$. We propose to use an estimate of the ideal cluster prototypes to update $\mathbf{W}$. Yet the computed $\{\mu_j^s\}$ and $\{\mu_j^t\}$ cannot be naively used for this purpose, not only because the correspondence between $\{\mathbf{w}_i\}$ and $\{\mu_j\}$ is unknown, but also the $k$-means result may contain very impure clusters, leading to non-representative prototypes. 

We  use the few-shot labeled data as well as samples with high-confident predictions to estimate the prototype for each class. Formally, we define $\mathcal{D}_s^{(i)} = \{\mathbf{x}|(\mathbf{x},y)\in\mathcal{D}_{s}, y=i\}$ and denote by $\mathcal{D}_{su}^{(i)}$ and $\mathcal{D}_{tu}^{(i)}$ the set of samples with high-confident label $i$ in source and target, respectively.
With $\mathbf{p(x)} = [\mathbf{p(x)}_1, \dots, \mathbf{p(x)}_{n_c}]$, 
$\mathcal{D}_{su}^{(i)} = \{\mathbf{x} | \mathbf{x}\in \mathcal{D}_{su}, \mathbf{p(x)}_{i} > t\}$, where $t$ is a confidence threshold; and similarly for $\mathcal{D}_{tu}^{(i)}$. Then the estimate of $\mathbf{w}_i$ from source and target domain can be computed as:
\begin{equation}
 \mathbf{\hat{w}}_i^s = \frac{1}{|\mathcal{D}_{s^+}^{(i)}|}\sum_{\mathbf{x}\in\mathcal{D}_{s^+}^{(i)}} \bm{V^s}(\mathbf{x});
 \mathbf{\hat{w}}_i^t = \frac{1}{|\mathcal{D}_{tu}^{(i)}|}\sum_{\mathbf{x}\in\mathcal{D}_{tu}^{(i)}} \bm{V^t}(\mathbf{x})
\end{equation}
where $\mathcal{D}_{s^+}^{(i)} = \mathcal{D}_{s}^{(i)} \cup \mathcal{D}_{su}^{(i)}$ and $\bm{V}(\mathbf{x})$ returns the representation in memory bank corresponding to $\mathbf{x}$. 

With only few labeled samples in source, it is hard to learn a representative prototype shared across domains. Instead of directly employing a global prototype for a class $i$, we further propose to update $\mathbf{w}_i$ in an domain adaptive manner, with $\mathbf{\hat{w}}_i^s$ during early training stage and with $\mathbf{\hat{w}}_i^t$ at later stage. This is because that $\mathbf{\hat{w}}_i^s$ is more robust in early training stage due to the few labeled source samples, while $\mathbf{\hat{w}}_i^t$ would be more representative later for target domain to get better adaptation performance. Specifically, we use $|\mathcal{D}_{tu}^{(i)}|$ to determine whether $\mathbf{\hat{w}}_i^t$ is robust to use:
\begin{align}
 \mathbf{w}_i = 
 \begin{cases}
 unit(\mathbf{\hat{w}}_i^s) & \text{if } |\mathcal{D}_{tu}^{(i)}| < t_w\\
 unit(\mathbf{\hat{w}}_i^t) & \text{otherwise}
 \end{cases} \label{eq:APCU}
\end{align}
where $unit(\cdot)$ normalizes the input vector and $t_w$ is a threshold hyper-parameter. 
\paragraph{Mutual Information Maximization}
In order for the above unified prototype-classifier learning paradigm to work well, the network is desired to have enough confident predictions, \textit{e.g.} $|\mathcal{D}^{(i)}| > t_w$, for all classes to get robust $\mathbf{\hat{w}}_i^s$ and $\mathbf{\hat{w}}_i^t$ for $i=1,\dots, n_c$. 
First, to promote the network to have diversified outputs over the dataset, we maximize the entropy of expected network prediction $\mathcal{H}(\mathbb{E}_{\mathbf{x}\in\mathcal{D}}[p(y|\mathbf{x}; \theta)])$, where $\theta$ denotes learnable parameters in both $F$ and $C$, and $\mathcal{D} = \mathcal{D}_{s}\cup \mathcal{D}_{su}\cup \mathcal{D}_{tu}$. Second, in order to get high-confident prediction for each sample, we leverage entropy minimization on the network output which has shown efficacy in label-scarce scenarios~\cite{grandvalet2005semi, berthelot2019mixmatch}. 
These two desired behaviors turn out to be equivalent to maximizing the mutual information between input and output: 
\begin{equation}
 \mathcal{I}(y;\mathbf{x}) = \mathcal{H}(\mathbf{p}_0) - \mathbb{E}_{\mathbf{x}}[\mathcal{H}(p(y|\mathbf{x};\theta))], 
\end{equation}
where the prior distribution $\mathbf{p}_0$ is given by $\mathbb{E}_\mathbf{x}[p(y|\mathbf{x};\theta)]$, and the detailed derivation is presented in the supplementary material. We can get the objective as:
\begin{equation}
 \mathcal{L}_\mathrm{MIM} = - \mathcal{I}(y;\mathbf{x})
\end{equation}

\begin{table*}[t]
\centering
\caption{Adaptation accuracy (\%) comparison on 1-shot and 3-shots per class on the Office dataset. }
\resizebox{0.9\textwidth}{!}{
\begin{threeparttable}
\begin{tabular}{l|c|c|c|c|c|c|c}
\toprule
\multirow{2}{*}{Method} & \multicolumn{7}{c}{Office: Target Acc. on 1-shot / 3-shots} \\ \cmidrule{2-8} 
 & A$\rightarrow$D & A$\rightarrow$W & D$\rightarrow$A & D$\rightarrow$W & W$\rightarrow$A & W$\rightarrow$D & Avg \\
\midrule
SO & 27.5 / 49.2 & 28.7 / 46.3 & 40.9 / 55.3 & 65.2 / 85.5 & 41.1 / 53.8 & 62.0 / 86.1 & 44.2 / 62.7 \\
MME~\cite{saito2019semi} & 21.5 / 51.0 & 12.2 / 54.6 & 23.1 / 60.2 & 60.9 / 89.7 & 14.0 / 52.3 & 62.4 / 91.4 & 32.3 / 66.5 \\
CDAN~\cite{long2018conditional}   & 11.2 / 43.7 & 6.2 / 50.1  & 9.1 / 65.1  & 54.8 / 91.6 & 10.4 / 57.0 & 41.6 / 89.8 & 22.2 / 66.2 \\
SPL~\cite{wang2020unsupervised} & 12.0 / 77.1 & 7.7 / 80.3 & 7.3 / 74.2 & 7.2 / 93.5 & 7.2 / 64.4 & 10.2 / 91.6 & 8.6 / 80.1 \\
CAN~\cite{kang2019contrastive} & 25.3 / 48.6 & 26.4 / 45.3 & 23.9 / 41.2 & 69.4 / 78.2 & 21.2 / 39.3 & 67.3 / 82.3 & 38.9 / 55.8 \\ 
MDDIA~\cite{jiang2020implicit} & 45.0 / 62.9 & 54.5 / 65.4 & 55.6 / 67.9 & 84.4 / 93.3 & 53.4 / 70.3 & 79.5 / 93.2 & 62.1 / 75.5 \\
CDS~\cite{kim2020cross} & 33.3 / 57.0 & 35.2 / 58.6 & 52.0 / 67.6 & 59.0 / 86.0 & 46.5 / 65.7 & 57.4 / 81.3 & 47.2 / 69.3 \\ \midrule
DANN + ENT~\cite{ganin2016domain}   & 32.5 / 57.6 & 37.2 / 54.1 & 36.9 / 54.1 & 70.1 / 87.4 & 43.0 / 51.4 & 58.8 / 89.4 & 46.4 / 65.7 \\
MME + ENT & 37.6 / 69.5 & 42.5 / 68.3 & 48.6 / 66.7 & 73.5 / 89.8 & 47.2 / 63.2 & 62.4 / 95.4 & 52.0 / 74.1 \\
CDAN + ENT   & 31.5 / 68.3 & 26.4 / 71.8 & 39.1 / 57.3 & 70.4 / 88.2 & 37.5 / 61.5 & 61.9 / 93.8 & 44.5 / 73.5 \\
CDS + ENT   & 40.4 / 61.2 & 44.7 / 66.7 & \underline{66.4} / 73.1 & 71.6 / 90.6 & 58.6 / 71.8 & 69.3 / 86.1 & 58.5 / 74.9 \\
CDS + MME + ENT & 39.4 / 61.6 & 43.6 / 66.3 & 66.0 / \underline{74.5} & 75.7 / 92.1 & 53.1 / \underline{73.0} & 70.9 / 90.6 & 58.5 / 76.3 \\
CDS + CDAN + ENT &  52.6 / 65.1  &  55.2 / 68.8  &  65.7 / 71.2  &  76.6 / 88.1  &  59.7 / 71.0  &  73.3 / 87.3  &  63.9 / 75.3 \\
CDS / MME + ENT$\textcolor{Violet}{^\dagger}$ & {\underline{55.4} / 75.7} & 57.2 / 77.2 & 62.8 / 69.7 & {\underline{84.9} / 92.1} & \underline{62.6} / 69.9 & \underline{77.7} / 95.4 & 65.3 / 80.0 \\
CDS / CDAN + ENT$\textcolor{Violet}{^\dagger}$   & 53.8 / \underline{78.1} & \underline{65.6} / \underline{79.8} & 59.5 / 70.7 & 83.0 / \underline{93.2} & 57.4 / 64.5 & 77.1 / \underline{97.4} & \underline{66.1} / \underline{80.6} \\
\midrule
PCS (Ours) & \textbf{60.2} / \textbf{78.2} & \textbf{69.8} / \textbf{82.9} & \textbf{76.1} / \textbf{76.4} & \textbf{90.6} / \textbf{94.1} & \textbf{71.2} / \textbf{76.3} & \textbf{91.8} / 96.0 & \textbf{76.6} / \textbf{84.0} \\
Improvement & \FG{+4.8} / \FG{+0.1} & \FG{+4.2} / \FG{+3.1} & \FG{+9.7} / \FG{+1.9} & \FG{+5.7} / \FG{+0.9} & \FG{+8.6} / \FG{+3.3} & \FG{+14.1} / -1.4 &  \FG{+10.5} / \FG{+3.4} \\
\bottomrule
\end{tabular}
\begin{tablenotes}
    \footnotesize
    \item[$\textcolor{Violet}{\dagger}$] Two-stage training results reported in~\cite{kim2020cross}.
\end{tablenotes}
\end{threeparttable}
}
\label{tab:office}
\end{table*}

\begin{table*}[]
\centering
\caption{Performance contribution of each part in PCS framework on Office. }
\resizebox{0.7\textwidth}{!}{%
\begin{tabular}{l|ccccccc}
\toprule
\multirow{2}{*}{Method} & \multicolumn{7}{c}{Office: Target Acc. on 1-shot / 3-shots} \\ \cmidrule{2-8} 
 & A$\rightarrow$D & A$\rightarrow$W & D$\rightarrow$A & D$\rightarrow$W & W$\rightarrow$A & W$\rightarrow$D & Avg \\
\midrule
$\mathcal{L}_{\mathrm{cls}}$ & 27.5 / 49.2 & 28.7 / 46.3 & 40.9 / 55.3 & 65.2 / 85.5 & 41.1 / 53.8 & 62.0 / 86.1 & 44.2 / 62.7 \\
$+\mathcal{L}_\mathrm{InSelf}$ & 39.0 / 55.6 & 38.6 / 55.1 & 47.2 / 68.5 & 71.7 / 89.4 & 50.9 / 68.4 & 65.1 / 90.6 & 52.1 / 71.3  \\
$+\mathcal{L}_\mathrm{CrossSelf}$ & 47.2 / 71.1 & 52.7 / 70.6 & 59.0 / 75.5 & 76.4 / 90.3 & 58.5 / 74.1 & 66.9 / 91.8 & 60.1 / 78.9 \\
$+\mathcal{L}_\mathrm{MIM}$ & 52.8 / 73.5 & 57.5 / 71.2 & 67.2 / 76.3 & 78.9 / 91.4 & 64.2 / 74.3 & 68.7 / 92.2 & 64.9 / 79.8  \\
\Gd $+$APCU (PCS) & \textbf{60.2} / \textbf{78.2} & \textbf{69.8} / \textbf{82.9} & \underline{76.1} / \textbf{76.4} & \textbf{90.6} / \textbf{94.1} & \textbf{71.2} / \textbf{76.3} & \textbf{91.8} / \textbf{96.0} & \textbf{76.6} / \textbf{84.0} \\ \hline
PCS w/o MIM & \underline{59.0} / \underline{75.9} & \underline{58.6} / \underline{76.5} & \textbf{76.2} / \textbf{76.4} & \underline{87.8} / \underline{93.2} & \underline{68.7} / \underline{74.7} & \underline{89.8} / \underline{95.0} & \underline{73.5} / \underline{82.0} \\
\bottomrule
\end{tabular}%
}
\label{tab:office_ablation}
\end{table*}

\subsection{PCS Learning for FUDA}
The PCS learning framework performs in-domain prototypical contrastive learning, cross-domain instance-prototype self-supervised learning, and unified adaptive prototype-classifier learning. Together with APCU in Eq. \ref{eq:APCU}, the overall learning objective is:
\begin{equation}
 \begin{aligned}
 \mathcal{L}_{PCS} = \ & \mathcal{L}_{\mathrm{cls}} + \lambda_\mathrm{in} \cdot \mathcal{L}_\mathrm{InSelf} \\
 & + \lambda_\mathrm{cross} \cdot \mathcal{L}_\mathrm{CrossSelf} + \lambda_\mathrm{mim} \cdot \mathcal{L}_\mathrm{MIM}\\
\end{aligned}
\end{equation}

\begin{table*}[]
\centering
\caption{Adaptation accuracy (\%) comparison on 3\% and 6\% labeled samples per class on the Office-Home dataset. }
\resizebox{0.97\textwidth}{!}{%
\begin{threeparttable}
\begin{tabular}{l|c|c|c|c|c|c|c|c|c|c|c|c|c}
\toprule[1.5pt]
\multirow{2}{*}{Method} & \multicolumn{13}{c}{Office-Home: Target Acc. (\%) } \\ \cmidrule{2-14} 
 & Ar $\rightarrow$Cl & Ar $\rightarrow$Pr & Ar $\rightarrow$Rw & Cl $\rightarrow$Ar & Cl $\rightarrow$Pr & Cl $\rightarrow$Rw & Pr $\rightarrow$Ar & Pr $\rightarrow$Cl & Pr $\rightarrow$Rw & Rw $\rightarrow$Ar & Rw $\rightarrow$Cl & Rw $\rightarrow$Pr & Avg \\
 \midrule
 \multicolumn{14}{c}{\textbf{3\% labeled source}} \\ \midrule
SO & 24.4 & 38.3 & 43.1 & 26.4 & 34.7 & 33.7 & 27.5 & 26.5 & 42.6 & 41.2 & 29.0 & 52.3 & 35.0 \\
MME~\cite{saito2019semi} & 4.5 & 15.4 & 25.0 & 28.7 & 34.1 & 37.0 & 25.6 & 25.4 & 44.9 & 39.3 & 29.0 & 52.0 & 30.1 \\
CDAN~\cite{long2018conditional} & 5.0 & 8.4 & 11.8 & 20.6 & 26.1 & 27.5 & 26.6 & 27.0 & 40.3 & 38.7 & 25.5 & 44.9 & 25.2 \\
MDDIA~\cite{jiang2020implicit} & 21.7 & 37.3 & 42.8 & 29.4 & 43.9 & 44.2 & 37.7 & 29.5 & 51.0 & 47.1 & 29.2 & 56.4 & 39.1 \\
CAN~\cite{kang2019contrastive} & 17.1 & 30.5 & 33.2 & 22.5 & 34.5 & 36.0 & 18.5 & 19.4 & 41.3 & 28.7 & 18.6 & 43.2 & 28.6 \\
CDS~\cite{kim2020cross} & 33.5 & 41.1 & 41.9 & 45.9 & 46.0 & 49.3 & 44.7 & 37.8 & 51.0 & 51.6 & 35.7 & 53.8 & 44.4 \\
\midrule
DANN + ENT~\cite{ganin2016domain} & 19.5 & 30.2 & 38.1 & 18.1 & 21.8 & 24.2 & 31.6 & 23.5 & 48.1 & 40.7 & 28.1 & 50.2 & 31.2 \\
MME + ENT & 31.2 & 35.2 & 40.2 & 37.3 & 39.5 & 37.4 & 48.7 & 42.9 & 60.9 & 59.3 & 46.4 & 58.6 & 44.8 \\
CDAN + ENT & 20.6 & 31.4 & 41.2 & 20.6 & 24.9 & 30.6 & 33.5 & 26.5 & 56.7 & 46.9 & 29.5 & 48.4 & 34.2 \\
CDS + ENT & 39.2 & 46.1 & 47.8 & 49.9 & 50.7 & 54.1 & 48.0 & 43.5 & 59.3 & 58.6 & 44.3 & 59.3 & 50.1 \\
CDS + MME + ENT & 39.4 & 48.0 & 52.1 & 50.0 & 52.3 & 56.4 & 47.8 & 44.2 & 60.6 & 61.1 & 45.3 & 62.1 & 51.6 \\
CDS + CDAN + ENT & \underline{43.8} & \underline{55.5} & \underline{60.2} & 51.5 & \underline{56.4} & \underline{59.6} & 51.3 & \underline{46.4} & 64.5 & 62.2 & \underline{52.4} & \underline{70.2} & \underline{56.2} \\
CDS / MME + ENT$\textcolor{Violet}{^\dagger}$ & 41.7 & 49.4 & 57.8 & \underline{51.8} & 52.3 & 55.9 & \underline{54.3} & 46.2 & \underline{69.0} & \underline{65.6} & 52.2 & 68.2 & 55.4 \\
CDS / CDAN + ENT$\textcolor{Violet}{^\dagger}$ & 37.7 & 49.2 & 56.5 & 49.8 & 51.9 & 55.9 & 50.0 & 42.3 & 68.1 & 63.1 & 48.7 & 67.5 & 53.4 \\
 \hline
 PCS (Ours) & \textbf{42.1} & \textbf{61.5} & \textbf{63.9} & \textbf{52.3} & \textbf{61.5} & \textbf{61.4} & \textbf{58.0} & \textbf{47.6} & \textbf{73.9} & \textbf{66.0} & \textbf{52.5} & \textbf{75.6} & \textbf{59.7} \\ 
Improvement & -1.7 & \FG{+6.0} & \FG{+6.1} & \FG{+3.7} & \FG{+5.1} & \FG{+1.8} & \FG{+3.7} & \FG{+1.2} & \FG{+4.9} & \FG{+0.4} & \FG{+0.1} & \FG{+5.4} & \FG{+3.5} \\ \hline \midrule

 \multicolumn{14}{c}{\textbf{6\% labeled source}} \\ \midrule
SO & 28.7 & 45.7 & 51.2 & 31.9 & 39.8 & 44.1 & 37.6 & 30.8 & 54.6 & 49.9 & 36.0 & 61.8 & 42.7 \\
MME~\cite{saito2019semi} & 27.6 & 43.2 & 49.5 & 41.1 & 46.6 & 49.5 & 43.7 & 30.5 & 61.3 & 54.9 & 37.3 & 66.8 & 46.0 \\
CDAN~\cite{long2018conditional} & 26.2 & 33.7 & 44.5 & 34.8 & 42.9 & 44.7 & 42.9 & 36.0 & 59.3 & 54.9 & 40.1 & 63.6 & 43.6 \\
MDDIA~\cite{jiang2020implicit} & 25.1 & 44.5 & 51.9 & 35.6 & 46.7 & 50.3 & 48.3 & 37.1 & 64.5 & 58.2 & 36.9 & 68.4 & 50.3 \\
CAN~\cite{kang2019contrastive} & 20.4 & 34.7 & 44.7 & 29.0 & 40.4 & 38.6 & 33.3 & 21.1 & 53.4 & 36.8 & 19.1 & 58.0 & 35.8 \\
CDS~\cite{kim2020cross} & 38.8 & 51.7 & 54.8 & 53.2 & 53.3 & 57.0 & 53.4 & 44.2 & 65.2 & 63.7 & 45.3 & 68.6 & 54.1 \\
\midrule
DANN + ENT~\cite{ganin2016domain} & 22.4 & 32.9 & 43.5 & 23.2 & 30.9 & 33.3 & 33.2 & 26.9 & 54.6 & 46.8 & 32.7 & 55.1 & 36.3 \\
MME + ENT & 37.2 & 42.4 & 50.9 & 46.1 & 46.6 & 49.1 & 53.5 & 45.6 & 67.2 & 63.4 & 48.1 & 71.2 & 51.8 \\
CDAN + ENT & 23.1 & 35.5 & 49.2 & 26.1 & 39.2 & 43.8 & 44.7 & 33.8 & 61.7 & 55.1 & 34.7 & 67.9 & 42.9 \\
CDS + ENT & 42.9 & 55.5 & 59.5 & 55.2 & 55.1 & 59.1 & 54.3 & 46.9 & 68.1 & 65.7 & 50.6 & 71.5 & 57.0 \\
CDS + MME + ENT & 41.7 & 58.1 & 61.7 & 55.7 & 56.2 & 61.3 & 54.6 & 47.3 & 68.6 & 66.4 & 50.3 & 72.1 & 57.8 \\
CDS + CDAN + ENT & \underline{45.4} & \underline{60.4} & \underline{65.5} & \underline{54.9} & \underline{59.2} & \underline{63.8} & 55.4 & \underline{49.0} & 71.6 & 66.6 & 54.1 & \underline{75.4} & \underline{60.1} \\
CDS / MME + ENT$\textcolor{Violet}{^\dagger}$ & 44.1 & 51.6 & 63.3 & 53.9 & 55.2 & 62.0 & 56.5 & 46.6 & 70.9 & \underline{67.7} & \underline{54.7} & 74.7 & 58.4 \\
CDS / CDAN + ENT$\textcolor{Violet}{^\dagger}$ & 39.0 & 51.3 & 63.1 & 51.0 & 55.0 & 63.6 & \underline{57.8} & 45.9 & \underline{72.8} & 65.8 & 50.4 & 73.5 & 57.4 \\
 \hline
PCS (Ours) & \textbf{46.1} & \textbf{65.7} & \textbf{69.2} & \textbf{57.1} & \textbf{64.7} & \textbf{66.2} & \textbf{61.4} & \textbf{47.9} & \textbf{75.2} & 67.0 & 53.9 & \textbf{76.6} & \textbf{62.6} \\
Improvement & \FG{+0.7} & \FG{+5.3} & \FG{+3.7} & \FG{+2.2} & \FG{+5.5} & \FG{+2.4} & \FG{+3.6} & -1.1 & \FG{+2.4} & -0.7 & -0.8 & \FG{+1.2} & \FG{+2.5} \\
\bottomrule
\end{tabular}%
\begin{tablenotes}
    \footnotesize
    \item[$\textcolor{Violet}{\dagger}$] Two-stage training results reported in~\cite{kim2020cross}.
\end{tablenotes}
\end{threeparttable}
}
\label{tab:officehome}
\vspace{-2mm}
\end{table*}

\begin{table}[]
\centering
\caption{Adaptation accuracy (\%) comparison on 0.1\% and 1\% labeled samples per class on the VisDA-2017 dataset. }
\vspace{-1mm}
\resizebox{0.8\columnwidth}{!}{%
\begin{threeparttable}
\begin{tabular}{l|cc}
\toprule[1.0pt]
\multirow{2}{*}{Method} & \multicolumn{2}{c}{VisDA: Target Acc. (\%)} \\ \cline{2-3}
 & 0.1\% Labeled & 1\% Labeled \\
\midrule
SO & 47.9 & 51.4 \\
MME~\cite{saito2019semi} & 55.6 & 69.4 \\
CDAN~\cite{long2018conditional} & 58.0 & 61.5 \\
MDDIA~\cite{jiang2020implicit} & 68.9 & 71.3 \\
CAN~\cite{kang2019contrastive} & 51.3 & 57.2 \\ 
CDS~\cite{kim2020cross} & 34.2 & 67.5 \\ \midrule
DANN + ENT~\cite{ganin2016domain} & 44.5 & 50.2 \\
MME + ENT & 54.0 & 66.1 \\
CDAN + ENT & 57.7 & 58.1 \\
CDS + ENT & 49.8 & 75.3 \\
CDS + ENT + MME & 60.0 & \underline{78.3} \\
CDS / MME + ENT$\textcolor{Violet}{^\dagger}$ & 62.5 & 69.4 \\
CDS / CDAN + ENT$\textcolor{Violet}{^\dagger}$ & \underline{69.0} & 69.1 \\ \midrule
PCS~(Ours) & \textbf{78.0} & \textbf{79.0} \\
Improvement & \FG{+9.0} & \FG{+0.7} \\
\bottomrule
\end{tabular}%
\begin{tablenotes}
    \footnotesize
    \item[$\textcolor{Violet}{\dagger}$] Two-stage training results reported in~\cite{kim2020cross}.
\end{tablenotes}
\end{threeparttable}
}
\label{tab:visda}
\vspace{-2mm}
\end{table}

\begin{table}[]
\centering
\caption{Adaptation accuracy (\%) comparison on 1-shot and 3-shots per class on the DomainNet dataset. }
\vspace{-1mm}
\resizebox{\columnwidth}{!}{%
\begin{tabular}{l|cccccccc}
\toprule[1.0pt]
\multirow{2}{*}{Method} & \multicolumn{8}{c}{DomainNet: Target Acc. (\%)} \\ \cline{2-9}
 & R$\veryshortarrow$C & R$\veryshortarrow$P & R$\veryshortarrow$S & P$\veryshortarrow$C & P$\veryshortarrow$R & C$\veryshortarrow$S & S$\veryshortarrow$P & Avg \\ \specialrule{0.5pt}{1pt}{1pt}
\multicolumn{9}{c}{\textbf{1-shot labeled source}} \\ \midrule
SO & 18.4 & 30.6 & 16.7 & 16.2 & \underline{28.9} & 12.7 & 10.5 & 19.1 \\
MME~\cite{saito2019semi} & 13.8 & 29.2 & 9.7 & 16.0 & 26.0 & 13.4 & 14.4 & 17.5 \\
CDAN~\cite{long2018conditional} & 16.0 & 25.7 & 12.9 & 12.6 & 19.5 & 7.2 & 8.0 & 14.6 \\
MDDIA~\cite{jiang2020implicit} & 18.0 & \underline{30.6} & 15.9 & 15.4 & 27.4 & 9.3 & 10.2 & 18.1 \\
CAN~\cite{kang2019contrastive} & 18.3 & 22.1 & 16.7 & 13.2 & 23.9 & 11.1 & 12.1 & 16.8 \\
\midrule
CDS~\cite{kim2020cross} & 16.7 & 24.4 & 11.1 & 14.1 & 15.9 & 13.4 & 19.0 & 16.4 \\
CDS + ENT & \underline{21.7} & 30.1 & \underline{18.2} & \underline{17.4} & 20.5 & 18.6 & \underline{22.7} & \underline{21.5} \\
CDS + MME + ENT & 21.2 & 28.8 & 15.5 & 15.8 & 17.6 & \underline{19.0} & 20.7 & 19.8 \\ \midrule
PCS (Ours) & \textbf{39.0} & \textbf{51.7} & \textbf{39.8} & \textbf{26.4} & \textbf{38.8} & \textbf{23.7} & \textbf{23.6} & \textbf{34.7} \\ 
Improvement & \FG{+17.3} & \FG{+21.1} & \FG{+21.6} & \FG{+9.0} & \FG{+9.9} & \FG{+4.7} & \FG{+0.9} & \FG{+13.2} \\
\hline \midrule
\multicolumn{9}{c}{\textbf{3-shots labeled source}} \\ \midrule
SO & 30.2 & 44.2 & 25.7 & 24.6 & 49.8 & 24.2 & 23.2 & 31.7 \\
MME~\cite{saito2019semi} & 22.8 & 46.5 & 14.5 & 25.1 & 50.0 & 20.1 & 24.9 & 29.1 \\
CDAN~\cite{long2018conditional} & 30.0 & 40.1 & 21.7 & 21.4 & 40.8 & 17.1 & 19.7 & 27.3 \\
MDDIA~\cite{jiang2020implicit} & 41.4 & 50.7 & 37.4 & 31.4 & \underline{52.9} & 23.1 & 24.1 & 37.3 \\
CAN~\cite{kang2019contrastive} & 28.1 & 33.5 & 25 & 24.7 & 46.9 & 23.3 & 20.1 & 28.8 \\ \midrule
CDS~\cite{kim2020cross} & 35.0 & 43.8 & 36.7 & 34.1 & 36.8 & 31.1 & 34.5 & 36.0 \\
CDS + ENT & \underline{44.5} & \underline{52.2} & 40.9 & \underline{40.0} & 47.2 & 33.0 & \underline{40.1} & \underline{42.5} \\
CDS + MME + ENT & 43.8 & 54.9 & \underline{41.1} & 38.9 & 45.9 & \underline{32.8} & 38.7 & 42.3 \\ \midrule
PCS (Ours) & \textbf{45.2} & \textbf{59.1} & \textbf{41.9} & \textbf{41.0} & \textbf{66.6} & 31.9 & 37.4 & \textbf{46.1} \\
Improvement & \FG{+0.7} & \FG{+6.9} & \FG{+0.8} & \FG{+1.0} & \FG{+13.7} & -0.9 & -2.7 & \FG{+3.6} \\
\bottomrule
\end{tabular}%
}
\label{tab:domainnet}
\vspace{-4mm}
\end{table}

\vspace{-4mm}
\section{Experiments}

\subsection{Experimental Setting}
\paragraph{Datasets.}
We evaluate our approach on four public datasets and choose labeled images in source domain based on previous work~\cite{kim2020cross}. \textbf{Office}~\cite{saenko2010adapting} is a real-world dataset for domain adaptation tasks. It contains 3 domains (Amazon, DSLR, Webcam) with 31 classes.
Experiments are conducted with 1-shot and 3-shots source labels per class in this dataset. 
\textbf{Office-Home}~\cite{venkateswara2017deep} is a more difficult dataset than Office, which consists of 4 domains (Art, Clipart, Product, Real) in 65 classes. 
Following~\cite{kim2020cross}, we look into the settings with 3\% and 6\% labeled source images per class, which means each class has 2 to 4 labeled images on average.
\textbf{VisDA-2017}~\cite{peng2017visda} is a challenging simulation-to-real dataset containing over 280K images across 12 classes. 
We validate our model on settings with 0.1\% and 1\% labeled source images per class as suggested in~\cite{kim2020cross}. 
\textbf{DomainNet}~\cite{peng2019moment} is a large-scale domain adaptation benchmark. Since some domains and classes are noisy, we follow~\cite{saito2019semi} and use a subset containing four domains (Clipart, Real, Painting, Sketch) with 126 classes.
We show results on settings with 1-shot and 3-shots source labels on this dataset.

\paragraph{Implementation Details.}
We use ResNet-101~\cite{he2016deep} (for DomainNet) and ResNet-50 (for other datasets) pre-trained on ImageNet~\cite{russakovsky2015imagenet} as our backbones. To enable a fair comparison with~\cite{kim2020cross}, we replaced the last FC layer with a 512-D randomly initialized linear layer. L2-normalizing are performed on the output features. We use $k$-means GPU implementation in faiss~\cite{faiss} for efficient clustering. We use SGD with momentum of 0.9, a learning rate of 0.01, a batch size of 64. More implementation details can be found in the supplementary material.

\subsection{Results on FUDA}

\paragraph{Baselines.}
\textbf{SO} is a model only trained using the labeled source images. \textbf{CDAN}~\cite{long2018conditional} and \textbf{MDDIA}~\cite{jiang2020implicit} are both state-of-the-art methods in UDA with a domain classifier to perform domain alignment. \textbf{MME}~\cite{saito2019semi} minimizes the conditional entropy of unlabeled target data with respect to the feature extractor and maximizes it with respect to the classifier. \textbf{CAN}~\cite{kang2019contrastive} uses clustering information to contrast discrepancy of source and target domain. \textbf{CDS}~\cite{kim2020cross} is a instance-based cross-domain self-supervised pre-training, which can be used for other domain adaptation methods and form \textit{two-stage} methods, such as \textbf{CDS / CDAN} and \textbf{CDS / MME}. We re-implement CDS into an end-to-end version by adding losses in two stage together and tuning the weight for different losses. We also investigate the \textit{one-stage} version of the methods above (\textbf{CDS + CDAN}, \textbf{CDS + MME}). Following~\cite{kim2020cross}, entropy minimization (\textbf{ENT}) on source is added to previous DA methods to obtain better baseline performance.

\begin{figure*}[!ht]
 \centering
 \includegraphics[width=6.6in, trim={0cm 0cm 0cm 0cm}]{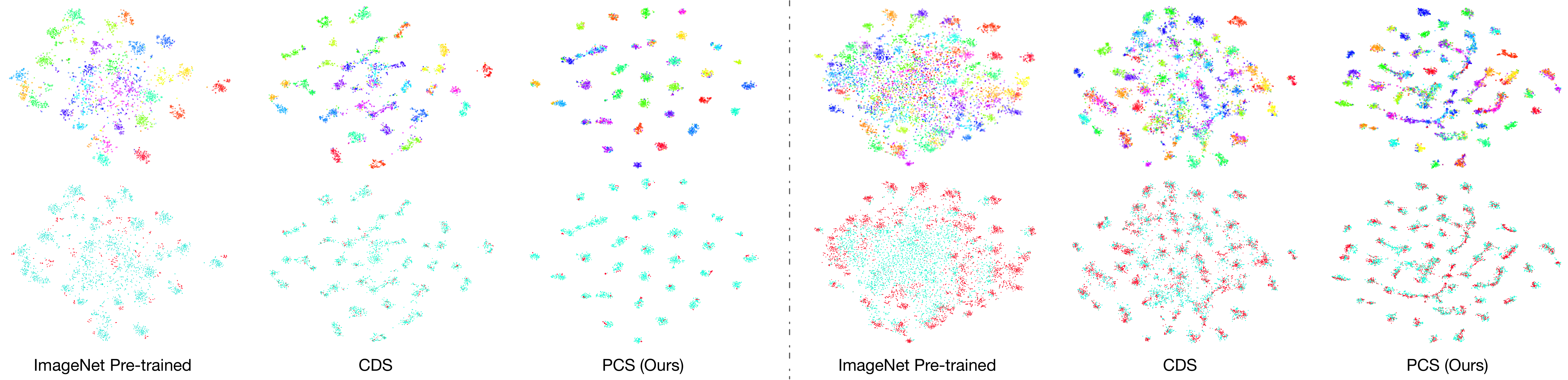}
 \caption{t-SNE visualization of ours and baselines on Office (left) and Office-Home (right). Top row: Coloring represents the class of each sample. Features with PCS are more discriminative than the ones with other methods. Bottom row: \textcolor{my_cyan}{Cyan} represents source features and \textcolor{myMaroon}{Red} represents target features. Feature from PCS are better-aligned between domains compared to other methods. 
 }
 \label{fig:tsne}
\end{figure*}

We compare the proposed PCS with state-of-the-art methods on FUDA (adaptation with few source labels). 
Extensive experiments are conducted on Office, Office-Home, VisDA-2017 and DomainNet, with the results presented in Table~\ref{tab:office}, \ref{tab:officehome}, \ref{tab:visda}, and \ref{tab:domainnet}, respectively. We can see that PCS outperforms the best state-of-the-arts in all the benchmarks, with large improvements: 10.5\%  and 3.4\% on Office, 4.3\% and 4.2\% on Office-Home, 9.0\% and 0.7\% on VisDA, 13.2\% and 3.6\% on DomainNet. If we look at the result of each domain pair in each dataset (\textit{e.g.} D $\rightarrow$ A in Office), PCS outperforms previous best in 47 out of 52 settings. Finally, as the number of labeled samples decreases, PCS shows larger performance improvements against the previous best methods, which demonstrates PCS is extremely beneficial in label-scarce adaptation scenarios.

\subsection{Ablation Study and Analysis}
Next, we investigate the effectiveness of each component in PCS on Office.
Table~\ref{tab:office_ablation} shows 
that adding each component contributes to the final results without any performance degradation. Comparing the last row in Table~\ref{tab:office_ablation} and Table~\ref{tab:office}, we can see even without MIM, PCS still outperforms all previous methods. 

We plot the learned features with t-SNE~\cite{maaten2008visualizing} on the DSLR-to-Amazon setting in Office, and Real-to-Clipart in Office-Home respectively in left and right of Figure~\ref{fig:tsne}. In the top row, the color represents the class of each sample; while in the bottom row, cyan represents source samples and red represents target samples. 
Compared to ImageNet pre-training and CDS, it qualitatively shows that PCS well clusters samples with the same class in the feature space; thus, PCS favors more discriminative features. Also, the features from PCS are more closely aggregated than ImageNet pre-training and CDS, which demonstrates that PCS learns a better semantic structure of the datasets. 

\begin{figure}[t]
 \centering
 \includegraphics[width=3.3in, trim={0.cm 0.cm 0 0}]{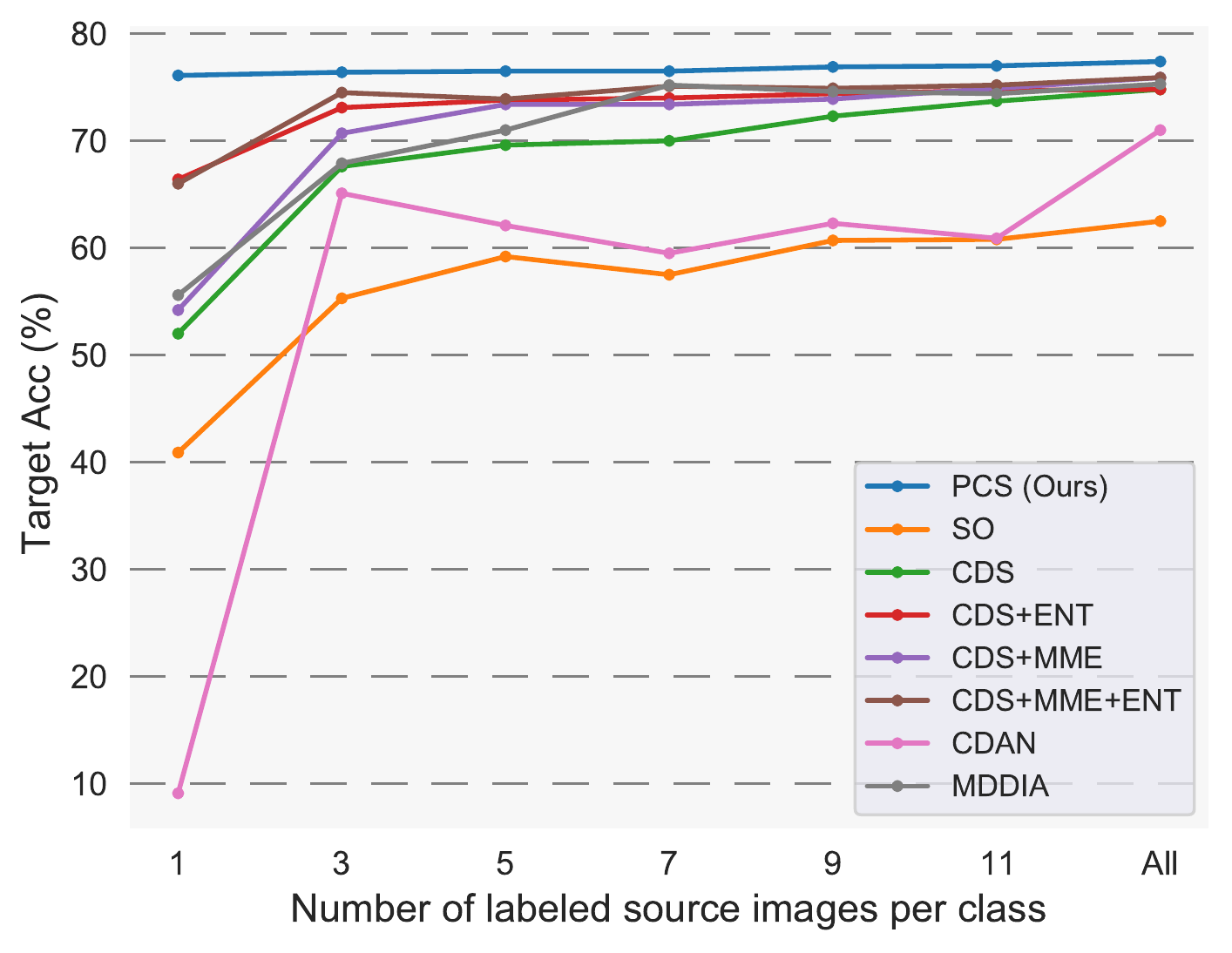}
 \caption{Sample efficiency comparison from DSLR to Amazon in Office dataset.}
 \label{fig:sample_efficiency_main}
\end{figure} 

\subsection{Sample Efficiency}

We compare our method with other state-of-the-art methods on Office dataset (DSLR as source and Amazon as target) with a varying number of source labels. From Figure~\ref{fig:sample_efficiency_main}, we can see that PCS outperforms all SOTA methods in all settings with different number of labeled samples. Moreover, our method is very label-efficient: 
a) For 1-shot image per class (31 labeled source images in total), PCS can achieve 76.1\% accuracy.
b) For the fully-labeled setting (498 labeled source images in total), PCS can achieve 77.4\% accuracy.
c) With \textbf{94\%} less labeled source images, the performance degradation of our method is only \textbf{1.3\%}. In short, \textit{with less labeled source data, PCS outperforms other works by a larger margin.}

\section{Conclusion}
In this paper, we investigated Few-shot Unsupervised Domain Adaptation where only few labeled samples are available in the source domain, and no labeled samples in the target domain. 
We proposed a novel Prototypical Cross-domain Self-supervised learning (PCS) framework that performs both in-domain and cross-domain prototypical self-supervised learning, as well as adaptive prototpe-classifier learning. 
We perform extensive experiments on multiple benchmark datasets, which demonstrates the superiority of PCS over previous best methods. PCS sets a new state of the art for Few-shot Unsupervised Domain Adaptation. 

{\small
\bibliographystyle{ieee_fullname}
\bibliography{z_egbib}
}

\clearpage
\appendix
\noindent\textbf{\Large Appendix}
\input{supp_backup}

\end{document}

%% file: supp_backup.tex
\begin{table*}[t]
\centering
\caption{Dataset statistics and labeled source used}
\label{tab:dataset}
\resizebox{0.8\textwidth}{!}{
\begin{tabular}{ccccc}
\toprule[1.3pt]
Dataset & \multicolumn{1}{c}{Domain} & \# total image & \# labeled images & \# classes \\ \midrule
\multirow{3}{*}{Office~\cite{saenko2010adapting}} & \multicolumn{1}{c}{Amazon (A)} & 2817 & \multirow{3}{*}{\begin{tabular}[c]{@{}c@{}}1-shot and 3-shots\\ labeled source\end{tabular}} & \multirow{3}{*}{31} \\ \cline{2-3}
 & \multicolumn{1}{c}{DSLR (D)} & 498 &  &  \\ \cline{2-3}
 & \multicolumn{1}{c}{Webcam (W)} & 795 &  &  \\ \midrule
\multirow{4}{*}{Office-Home~\cite{venkateswara2017deep}} & \multicolumn{1}{c}{Art (Ar)} & 2427 & \multirow{4}{*}{\begin{tabular}[c]{@{}c@{}}3\% and 6\%\\ labeled source\end{tabular}} & \multirow{4}{*}{65} \\ \cline{2-3}
 & \multicolumn{1}{c}{Clipart (Cl)} & 4365 &  &  \\ \cline{2-3}
 & \multicolumn{1}{c}{Product (Pr)} & 4439 &  &  \\ \cline{2-3}
 & \multicolumn{1}{c}{Real (Rw)} & 4357 &  &  \\ \midrule
\multirow{2}{*}{VisDA~\cite{peng2017visda}} & \multicolumn{1}{c}{Synthetic (Syn)} & 152K & \multirow{2}{*}{\begin{tabular}[c]{@{}c@{}}0.1\% and 1\%\\ labeled source\end{tabular}} & \multirow{2}{*}{12} \\ \cline{2-3}
 & \multicolumn{1}{c}{Real (Rw)} & 55K &  &  \\ \midrule
\multirow{4}{*}{DomainNet~\cite{peng2019moment}} & \multicolumn{1}{c}{Clipart (C)} & 18703 & \multirow{4}{*}{\begin{tabular}[c]{@{}c@{}}1-shot and 3-shots\\ labeled source\end{tabular}} & \multirow{4}{*}{126} \\ \cline{2-3}
 & \multicolumn{1}{c}{Painting (P)} & 31502 &  &  \\ \cline{2-3}
 & \multicolumn{1}{c}{Real (R)} & 70358 &  &  \\ \cline{2-3}
 & \multicolumn{1}{c}{Sketch (S)} & 24582 &  &  \\ \bottomrule
\end{tabular}
}
\end{table*}

\section{Proof of Equation (13)
}


As mentioned in Section 3.2 of the main paper, in order for the prototype-classifier learning paradigm to work well, the network is desired to have enough confident predictions for all classes to get robust $\mathbf{\hat{w}}_i^s$ and $\mathbf{\hat{w}}_i^t$. First, to promote the network to have diversified outputs, we propose to maximize the entropy of expected network prediction $\mathcal{H}(\mathbb{E}_{\mathbf{x}}[p(y|\mathbf{x}; \theta)])$. Second, to get high-confident prediction for each sample, we perform entropy minimization on the network output. So the overall objective is: 
\begin{equation}
\max \mathcal{H}(\mathbb{E}_{\mathbf{x}}[p(y|\mathbf{x}; \theta)]) - \mathbb{E}_{\mathbf{x}}[\mathcal{H}(p(y|\mathbf{x};\theta))].
\end{equation}
Now we show that this objective equals maximizing the mutual information between input and output, \textit{i.e.} $\mathcal{I}(y; \mathbf{x})$:
{\allowdisplaybreaks
\begin{align}
    &\mathcal{H}( \mathbb{E}_\mathbf{x} [p(y | \mathbf{x}; \theta)]) - \mathbb{E}_\mathbf{x}[ \mathcal{H}( p(y | \mathbf{x}; \theta))] \\
    =&\mathbb{E}_\mathbf{x}\left[ \sum_{i = 1}^L p(y_i | \mathbf{x}) \log p(y_i | \mathbf{x}) \right] \notag \\
        & \quad \quad - \sum_{i = 1}^L \mathbb{E}_\mathbf{x}[ p(y_i | \mathbf{x}) ] \log \mathbb{E}_\mathbf{x}[ p(y_i | \mathbf{x})] ] \\
    =&\mathbb{E}_\mathbf{x}\left[ \sum_{i = 1}^L p(y_i | \mathbf{x}) \log p(y_i | \mathbf{x}) \right] \notag \\
       & \quad \quad - \mathbb{E}_\mathbf{x} \left[ \sum_{i = 1}^L p(y_i | \mathbf{x}) \log \mathbb{E}_\mathbf{x}[p(y_i | \mathbf{x})] \right] \\
    = &\mathbb{E}_\mathbf{x}\left[ \sum_{i = 1}^L p(y_i | \mathbf{x}) \log \frac{p(y_i | \mathbf{x})}{\mathbb{E}_\mathbf{x}[p(y_i | \mathbf{x})]}\right] \\
    =&\mathbb{E}_\mathbf{x}\left[ \int p(y | \mathbf{x}) \log \frac{p(y | \mathbf{x})}{\mathbb{E}_\mathbf{x}[p(y | \mathbf{x})]} \dd{y} \right] \\
    =& \int p(\mathbf{x}) \dd{\mathbf{x}} \int p(y | \mathbf{x}) \log \frac{p(y | \mathbf{x})}{\int p(\mathbf{x}) p(y | \mathbf{x}) \dd{\mathbf{x}}} \dd{y} \\
    =& \int p(\mathbf{x}) \dd{\mathbf{x}} \int p(y | \mathbf{x}) \log \frac{p(y | \mathbf{x})}{p(y)} \dd{y} \\
    =& \iint p(y, \mathbf{x}) \log \frac{p(y, \mathbf{x})}{p(y)p(\mathbf{x})} \dd{y} \dd{\mathbf{x}} ~~=~~\mathcal{I}(y; \mathbf{x})
\end{align}
}
In addition, we estimate $\mathcal{H}(\mathbb{E}_{\mathbf{x}}[p(y|\mathbf{x}; \theta)])$ with $\sum_{x\in \mathcal{D}} p(y|\mathbf{x}; \theta)\log \mathbf{\hat p}_0$, where $\mathbf{\hat p}_0$ is a moving average of $p(y|\mathbf{x}; \theta)$.

\section{Additional Datasets Details}

Overall statistics of the datasets and the number of labeled source examples used in our experiments can be found in Table~\ref{tab:dataset}. For Office~\cite{saenko2010adapting}, Office-Home~\cite{venkateswara2017deep} and VisDA~\cite{peng2017visda}, we follow the same setting in~\cite{kim2020cross}, randomly sampling labeled images from the source domain and ensure that each class has at least one labeled example. 
For DomainNet~\cite{peng2019moment}, we use the same split files as~\cite{saito2019semi} and further select 1-shot and 3-shots labeled samples in the training set for each class.

\section{Additional Implementation Details}

We implemented our model in PyTorch~\cite{paszke2019pytorch}. We choose batch size of  64 for both source and target in self-supervised learning and batch size of 32 for the classification loss. The learning rate ratio between linear layer and convolution layer is set to $1:0.1$. We use SGD with weight decay rate $5e^{-4}$. For Office and Office-Home, we adaptively set temperature $\phi$ according to~\cite{li2020prototypical}. For VisDA and DomainNet, we fix $\phi$ to be 0.1 for more stable training. We set temperature $\tau$ to be 0.1 in all experiments. 
We choose hyper-parameters $\lambda_{\mathrm{in}}$ and $\lambda_{\mathrm{cross}}\in \{1, 0.5\}$, and the weight $\lambda_{\mathrm{mim}}\in \{0.05, 0.01\}$.
As for parameters $m$ (momentum for memory bank update) and $M$ (number of $k$-means in $\mathcal{L}_\mathrm{InSelf}$), we set $m=0.5$ and $M=20$. 

We use spherical $k$-means for clustering and set half of the number of clusters in $k$-means to be the number of the classes $n_c$, and the rest to be $2n_c$. We compute the weight for cosine classifier only using source images for the first 5 epochs and set $t_w$ to be around half of the average number of images per class. New prototypes (\textit{i.e.} centroids of clusters and weights of cosine classifier) are computed per epoch for both self-supervised learning and classification. 





\begin{table*}[th]
\caption{Averaged accuracy and standard deviation of PCS on three runs of 1-shot and 3-shots on Office dataset.}
\centering
\begin{tabular}{@{}ccccccc@{}}
\toprule
Labeled Source & A$\rightarrow$D & A$\rightarrow$W & D$\rightarrow$A & D$\rightarrow$W & W$\rightarrow$A & W$\rightarrow$D \\ \midrule
1-shot & 60.2$\pm$1.9 & 69.8$\pm$0.8 & 76.1$\pm$0.4 & 90.6$\pm$0.8 & 71.2$\pm$1.0 & 91.8$\pm$1.9 \\
3-shots & 78.2$\pm$1.8 & 82.9$\pm$1.1 & 76.4$\pm$0.5 & 94.1$\pm$0.1 & 76.3$\pm$0.7 & 96.0$\pm$0.7 \\ \bottomrule
\end{tabular}
\label{tab:multirun}
\end{table*}

\begin{figure*}[t]
 \centering
 \hspace*{0.1cm}
 \begin{subfigure}{.5\textwidth}
    \centering
    \hspace*{-1cm}\includegraphics[width=\linewidth, trim={0 0 1.5cm 1cm}, clip]{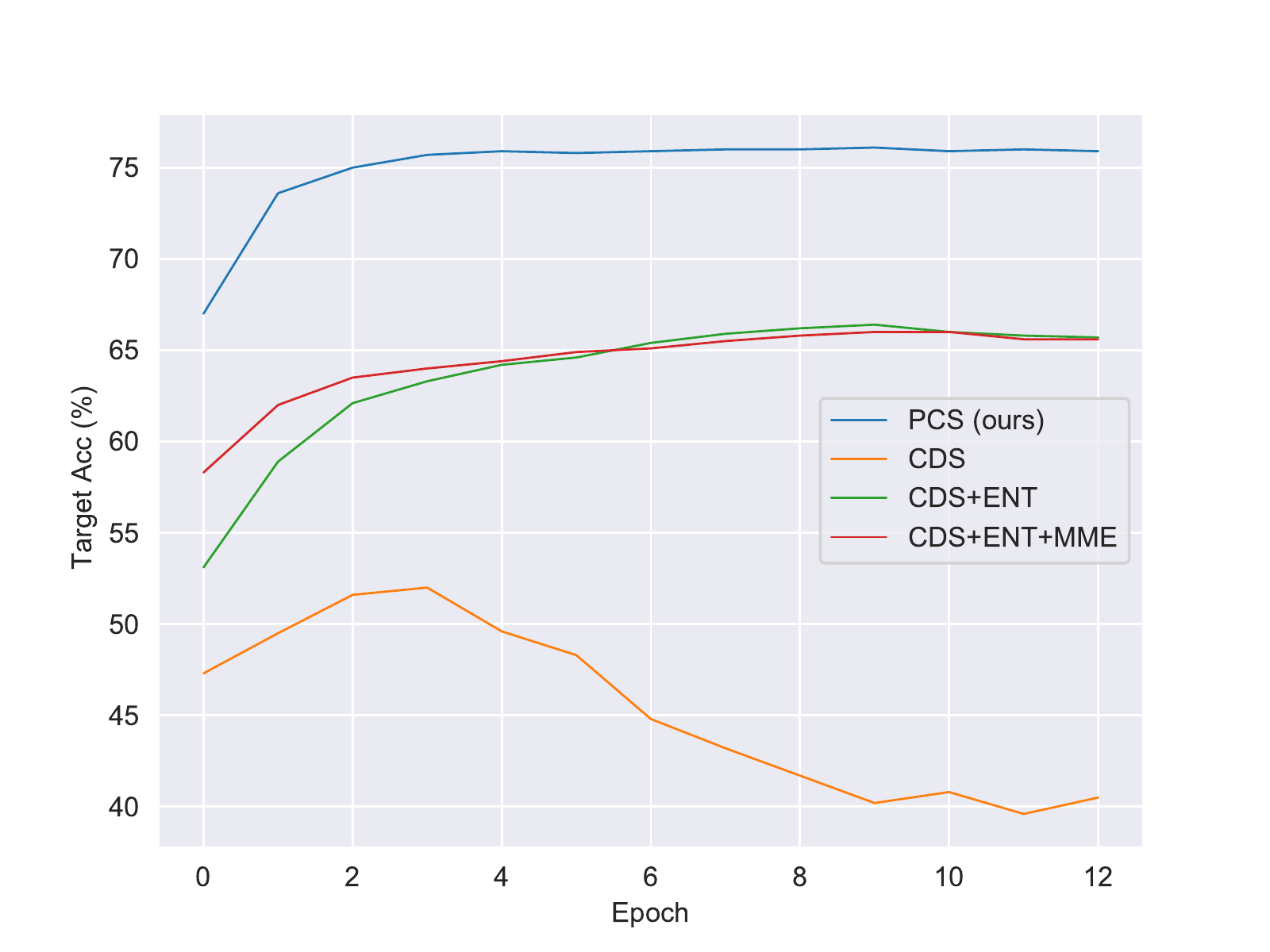}
    \caption{Target Acc. with cosine classifier vs. training epochs}
    \label{fig:accuracy_epoch:1}
 \end{subfigure}%
 \begin{subfigure}{.5\textwidth}
    \centering
    \hspace*{-1cm}\includegraphics[width=\linewidth, trim={0 0 1.5cm 1cm}, clip]{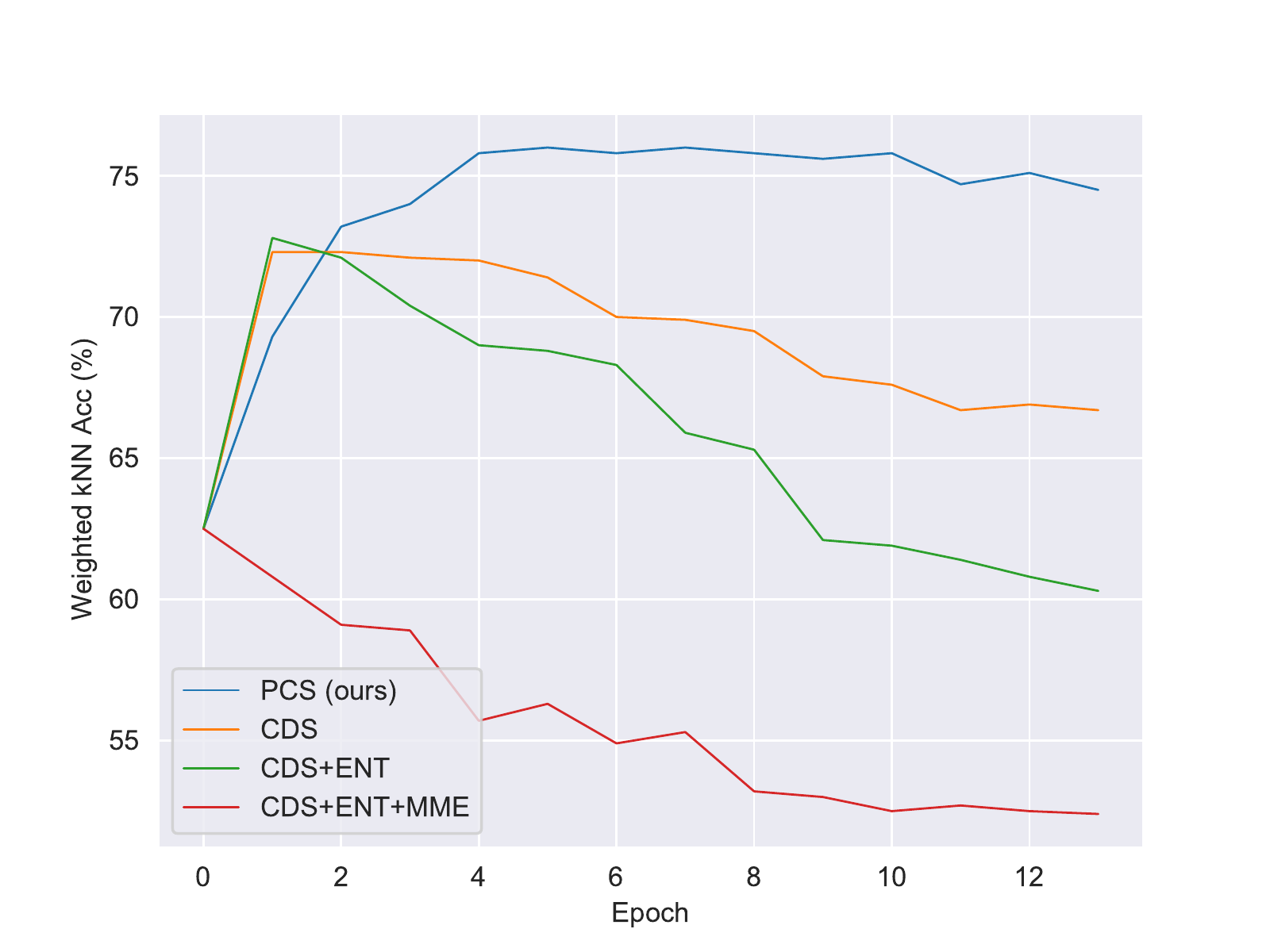}
    \caption{Target Acc. with Weighted kNN vs. training epochs}
    \label{fig:accuracy_epoch:2}
 \end{subfigure}%
 \caption{Stability of Target Accuracy during training procedure. }
 \label{fig:accuracy_epoch}
\end{figure*}

\section{Stability Analysis of PCS}

To show the performance stability of PCS, we conduct multiple runs with three different random seeds. Table~\ref{tab:multirun} reports the averaged accuracy and standard deviation of the three runs on the 1-shot and 3-shots settings of Office.

Figure~\ref{fig:accuracy_epoch} shows adaptation accuracy vs. training epochs using cosine classifier (Figure~\ref{fig:accuracy_epoch:1}) and weighted kNN classifier (Figure~\ref{fig:accuracy_epoch:2}). From the plots, we have the following observations. (1) The target accuracy of PCS increases more steadily and robustly compared to other methods. In Figure~\ref{fig:accuracy_epoch:1}, CDS starts decreasing at Epoch 3. In Figure~\ref{fig:accuracy_epoch:2}, CDS and CDS+ENT starts decreasing at Epoch 1; while CDS+ENT+MME decreases from the beginning of training. In contrast, the performance of PCS increases smoothly until the end of training. (2) PCS converges much faster than other methods. We can see in Figure~\ref{fig:accuracy_epoch:1} that PCS plateaus at around Epoch 3, while CDS+ENT and CDS+ENT+MME reaches the best performance at Epoch 9 and 10.

\begin{figure*}[t]
    \centering
    \begin{subfigure}{\textwidth}
        \centering
        \includegraphics[width=6in]{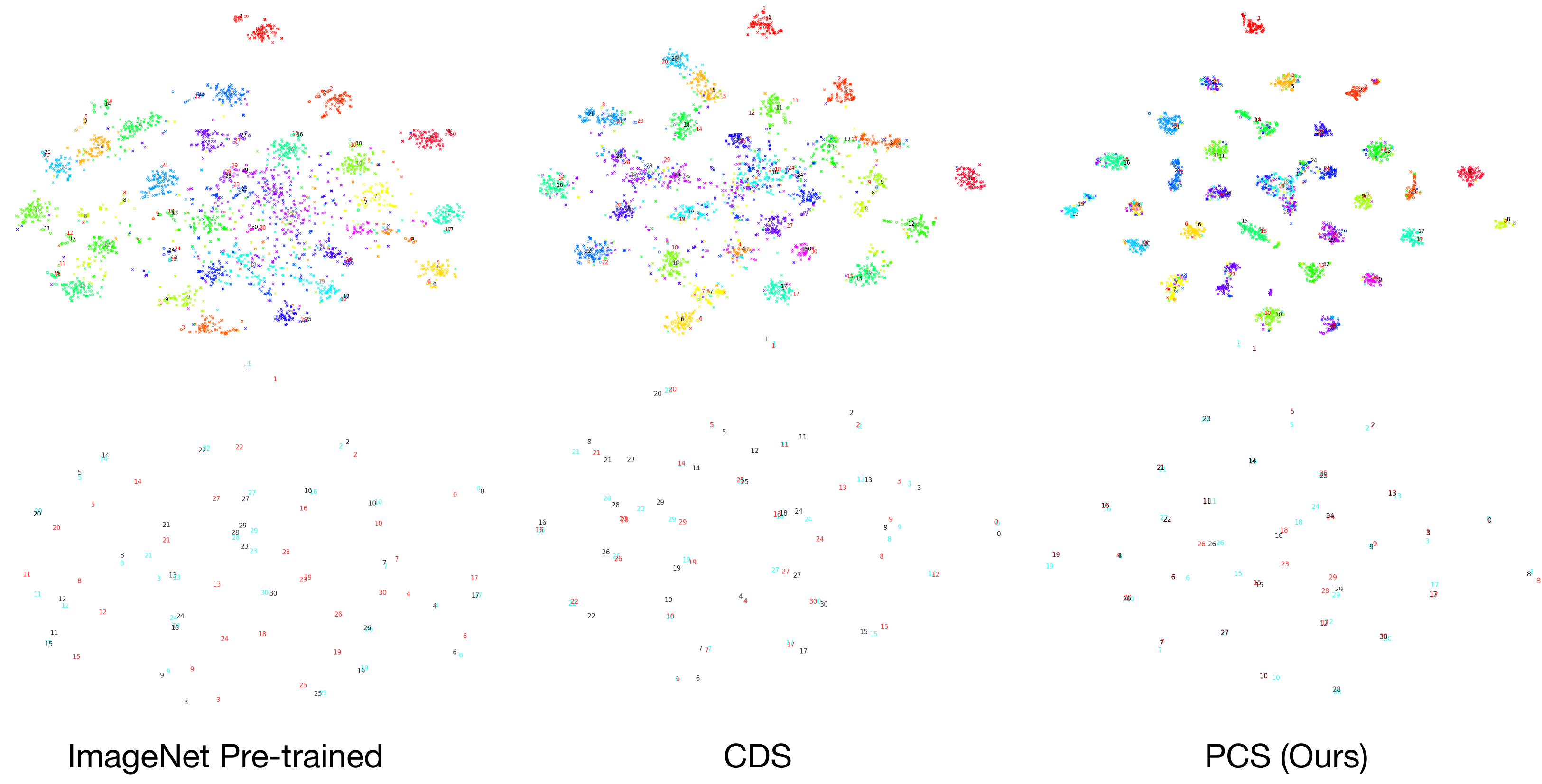}
        \caption{Office (D$\rightarrow$A with 1-shot labeled source per class)}
        \label{fig:tsne_office_proto}
    \end{subfigure}%
    \par\bigskip
    
    \begin{subfigure}{\textwidth}
        \centering
        \includegraphics[width=6in]{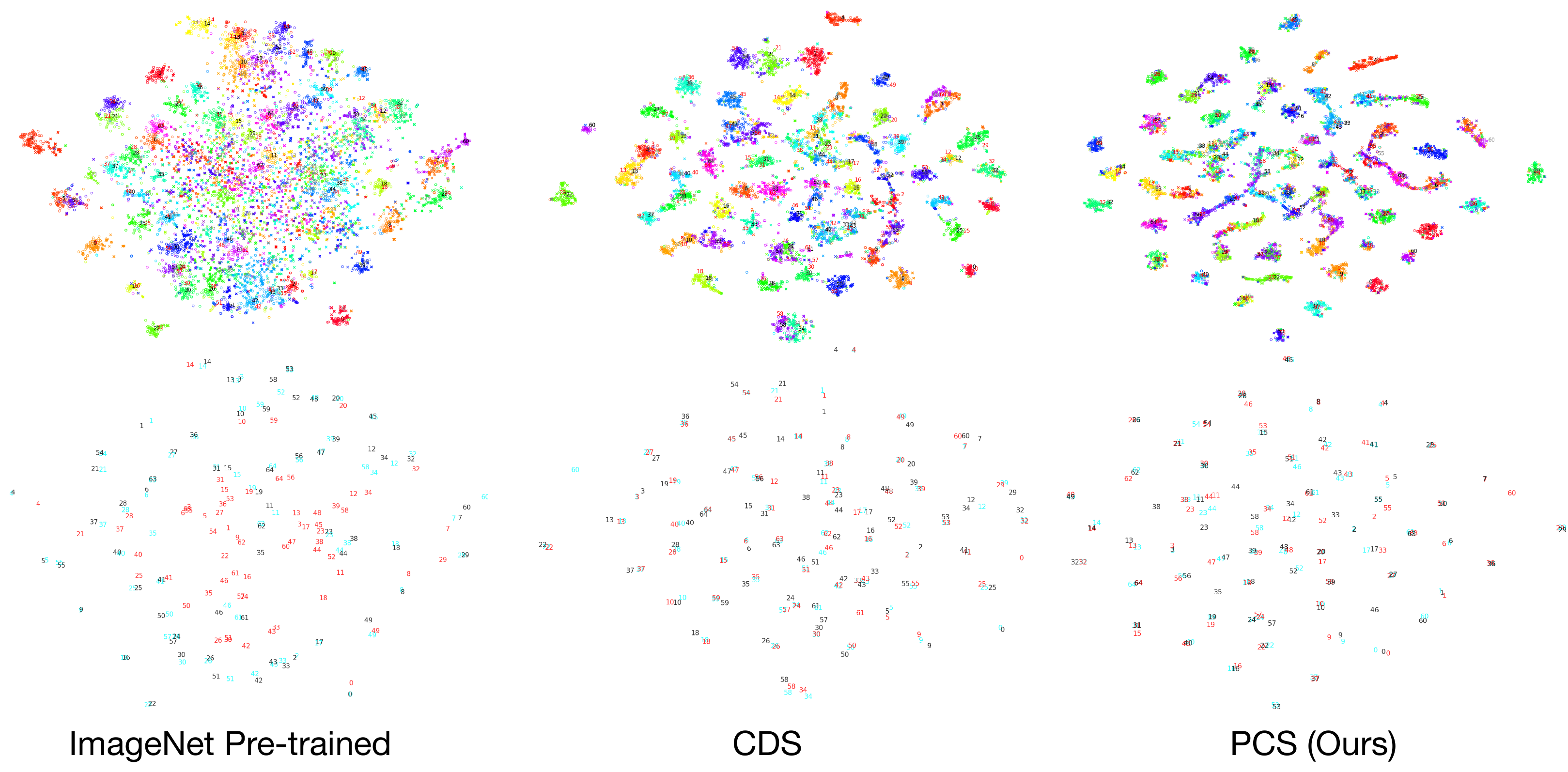}
        \caption{Office-Home (Rw$\rightarrow$Cl with 3\% labeled source per class)}
        \label{fig:tsne_officehome_proto}
    \end{subfigure}%
    \caption{t-SNE visualization of ours and baselines on Office (a) and Office-Home (b). Top row: Coloring represents the class of each sample, and shape represents domain (circle for source and cross for target). Features with PCS are more discriminative than the ones with other methods. Bottom row: each number represents a centroid for corresponding class. \textbf{\textcolor{my_cyan}{Cyan}} represents centroids of source images based on ground truth and \textbf{\textcolor{myMaroon}{Red}} for target. \textbf{\textcolor{black}{Black}} represents prototypes of the classifier. Centroids from PCS are better-aligned between domains compared to other methods. (Zoom in for more details).}
    \label{fig:tsne_proto}
\end{figure*}%

\section{Quantitative Feature Analysis}

To quantitatively compare the quality of learned features with different approaches, we perform classification with weighted $k$-nearest neighbor (kNN) classifier proposed by Wu~\etal~\cite{wu2018unsupervised} in a cross-domain manner. Specifically, given a test image $\mathbf{x}^t$, we first compute its normalized feature $\mathbf{f}^t = F(\mathbf{x}^t)$, and then compare it again embeddings of all source images in the source memory bank \bm{$V^s$} using cosine similarity $s_{i}=\mathrm{cos}(\mathbf{f}^t, \mathbf{v}_i^s)$. The top $k$ nearest neighbors in the source domain, $\mathcal{N}_k$, would be used to make the final prediction with weighted voting. Specifically, class $c$ would get weight $w_c = \sum_{i\in\mathcal{N}_k} \alpha_i \cdot \mathrm{1}(c_i=c)$, in which $\alpha_i$ is the contributing weight of neighbor $\mathbf{v}_i^s$ defined as $\alpha_i = \exp(s_{i} / \tau)$. We set $\tau=0.07$ and $k=200$ as in~\cite{wu2018unsupervised}.

\begin{table}[]
\centering
\caption{Accuracy of cross-domain weighted kNN with different SSL methods.}
\begin{tabular}{lcc}
\toprule
Method & D$\rightarrow$A & Rw$\rightarrow$Cl \\ \midrule
ImageNet pre-train & 62.5 & 40.6 \\
ID~\cite{wu2018unsupervised} & 70.3 & 51.9 \\
CDS~\cite{kim2020cross} & 72.5 & 53.7 \\
protoNCE~\cite{li2020prototypical} & 72.3 & 49.3 \\
\Gd $\mathcal{L}_\mathrm{InSelf} + \mathcal{L}_\mathrm{CrossSelf}$ & \textbf{75.5} & \textbf{55.3} \\ \midrule
\end{tabular}
\label{tab:ssl_feature_analysis}
\end{table}

\begin{table}[t]
\centering
\caption{Accuracy of cross-domain weighted kNN with different FUDA methods.}
\begin{tabular}{lcc}
\toprule
Method & D$\rightarrow$A (1-shot) & Rw$\rightarrow$Cl (3\%) \\ \midrule
CDS~\cite{kim2020cross} & 72.3 & 57.6 \\
CDS + ENT & 72.8 & 58.6 \\
CDS + MME + ENT & 60.8 & 59.2 \\
\Gd PCS (Ours) & \textbf{76.0} & \textbf{59.3} \\ \bottomrule
\end{tabular}
\label{tab:feature_analysis}
\end{table}

We perform the above cross-domain kNN classification on models trained with 1) only cross-domain self-supervised learning methods, and 2) Few-shot Unsupervised Domain Adaptation methods, with the results shown in Table~\ref{tab:ssl_feature_analysis} and Table~\ref{tab:feature_analysis}, respectively. 
From the results, we can see that both the proposed cross-domain prototypical self-supervised learning method and the whole PCS framework outperforms previous approaches. 

\begin{figure*}[]
 \centering
 \includegraphics[width=6.85in, trim={0 0 0 0}]{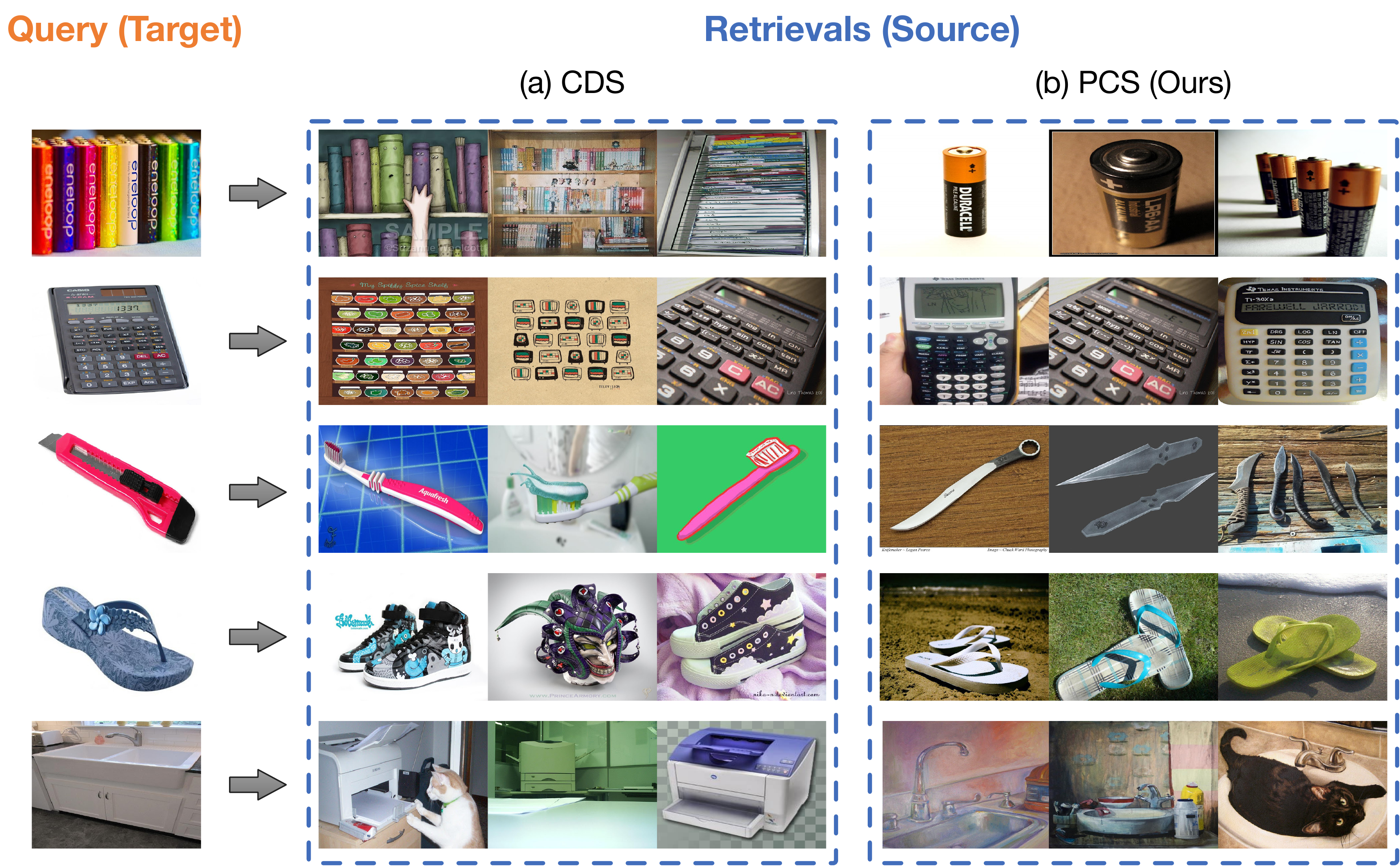}
 \caption{Image retrieval examples of the closest cross-domain neighbors using CDS (a) and PCS (b) in Office-Home (Target: Real, Source: Art).}
 \label{fig:retrieval}
\end{figure*}

\begin{table}[t]
\centering
\caption{Sum of pair-wise cosine-similarity between prototypes in Office and Office-Home. }
\begin{tabular}{lcc}
\toprule
Method & D$\rightarrow$A (1-shot) & Rw$\rightarrow$Pr (3\%) \\ \midrule
SO & 0.44 & -0.71 \\
CDS~\cite{kim2020cross} & 0.43 & -0.71 \\
PCS w/o APCU & -53.3 & -22.8 \\
\Gd PCS (Ours) & \textbf{-58.4} & \textbf{-26.5} \\ \bottomrule
\end{tabular}
\label{tab:sum_distance}
\end{table}

\section{Prototype Quality Comparison}
To further compare how well source and target are aligned, we provide more t-SNE~\cite{maaten2008visualizing} visualizations on Office (D$\rightarrow$A) and Office-Home (Rw$\rightarrow$Cl) in Figure~\ref{fig:tsne_office_proto} and \ref{fig:tsne_officehome_proto}, comparing ImageNet Pre-training, CDS~\cite{kim2020cross} and PCS. Specifically, we plot representations for all samples (top in both Figures), as well as the prototypes (normalized average representation) for each class. In top rows of both figures, the color of a sample represents its class, and samples from different domains are represented by different shapes (circles for source and crosses for target, best view after zooming in). In bottom rows of both figures, the number of a prototype represents its class index, and color represent the domain of the prototype (Cyan for source, Red for target, and Black for prototype weight of the classifier). As we can see from Figure~\ref{fig:tsne_proto}, for each class, the prototypes of source, target and the weight vector of classifier get more aggregated with PCS than other methods, which demonstrates that PCS could better align source and target representations for each category.

In a well-learned feature embedding space, prototypes of different classes should be far / different from each other. To quantitatively measure the similarity of the learned prototypes, we compute the sum of cosine similarities between all pairs of prototypes. From the results shown in Table~\ref{tab:sum_distance}, we can see that the prototypes learned with PCS have the least similarities, indicating that PCS learns an embedding space with better semantic structure.

\begin{table*}[]
\centering
\caption{Performance contribution of each part in PCS framework on Office-Home. }
\resizebox{0.96\textwidth}{!}{%
\begin{tabular}{l|c|c|c|c|c|c|c|c|c|c|c|c|c}
\toprule[1.3pt]
\multirow{2}{*}{Method} & \multicolumn{13}{c}{Office-Home: Target Acc.} \\ \cmidrule{2-14} 
 & Ar $\rightarrow$Cl & Ar $\rightarrow$Pr & Ar $\rightarrow$Rw & Cl $\rightarrow$Ar & Cl $\rightarrow$Pr & Cl $\rightarrow$Rw & Pr $\rightarrow$Ar & Pr $\rightarrow$Cl & Pr $\rightarrow$Rw & Rw $\rightarrow$Ar & Rw $\rightarrow$Cl & Rw $\rightarrow$Pr & Avg \\
 \midrule
 \multicolumn{14}{c}{\textbf{3\% labeled source}} \\ \midrule
 $\mathcal{L}_{\mathrm{cls}}$ & 24.4 & 38.3 & 43.1 & 26.4 & 34.7 & 33.7 & 27.5 & 26.5 & 42.6 & 41.2 & 29.0 & 52.3 & 35.0 \\
$+\mathcal{L}_\mathrm{InSelf}$ & 34.6 & 48.3 & 54.7 & 49.2 & 53.1 & 57.1 & 48.2 & 40.6 & 62.9 & 57.9 & 44.9 & 68.8 & 51.7 \\
$+\mathcal{L}_\mathrm{CrossSelf}$ & 36.5 & 53.7 & 56.6 & 51.2 & 57.9 & 58.8 & 51.2 & 42.8 & 66.2 & 61.5 & 50.1 & 72.2 & 54.9 \\
$+\mathcal{L}_\mathrm{MIM}$ & 37.2 & 55.9 & 58.8 & 51.5 & 59.4 & 59.0 & 53.2 & 43.0 & 68.2 & 62.0 & 50.2 & 72.5 & 55.9 \\
\Gd $+$APCU (PCS) & \textbf{42.1} & \textbf{61.5} & \textbf{63.9} & \textbf{52.3} & \textbf{61.5} & \textbf{61.4} & \textbf{58.0} & \textbf{47.6} & \textbf{73.9} & \textbf{66.0} & \textbf{52.5} & \textbf{75.6} & \textbf{59.7} \\ \hline \midrule
 \multicolumn{14}{c}{\textbf{6\% labeled source}} \\ \midrule
$\mathcal{L}_{\mathrm{cls}}$ & 28.7 & 45.7 & 51.2 & 31.9 & 39.8 & 44.1 & 37.6 & 30.8 & 54.6 & 49.9 & 36.0 & 61.8 & 42.7 \\
$+\mathcal{L}_\mathrm{InSelf}$ & 40.8 & 57.6 & 65.5 & 54.5 & 62.4 & 62.7 & 54.6 & 43.1 & 73.6 & 64.2 & 44.7 & 75.9 & 58.3 \\
$+\mathcal{L}_\mathrm{CrossSelf}$ & 40.8 & 59.5 & 66.9 & 55.5 & 64.1 & 63.1 & 57.2 & 46.2 & 73.9 & 65.0 & 52.0 & 76.9 & 60.1 \\
$+\mathcal{L}_\mathrm{MIM}$ & 42.1 & 60.2 & 68.5 & 55.9 & 64.4 & 63.5 & 59.1 & 47.1 & 74.4 & 66.6 & 52.1 & 77.0 & 60.9 \\
\Gd $+$APCU (PCS) & \textbf{46.1} & \textbf{65.7} & \textbf{69.2} & \textbf{57.1} & \textbf{64.7} & \textbf{66.2} & \textbf{61.4} & \textbf{47.9} & \textbf{75.2} & \textbf{67.0} & \textbf{53.9} & \textbf{76.6} & \textbf{62.6} \\ \bottomrule
\end{tabular}%
}
\label{tab:officehome_ablation}
\end{table*}

\begin{table*}[]
\centering
\caption{Adaptation accuracy (\%) comparison on fully-labeled setting on the Office-Home dataset.}
\resizebox{\linewidth}{!}{%
\begin{tabular}{lccccccccccccc}
\toprule
Method & Ar$\rightarrow$Cl & Ar$\rightarrow$Pr & Ar$\rightarrow$Rw & Cl$\rightarrow$Ar & Cl$\rightarrow$Pr & Cl$\rightarrow$Rw & Pr$\rightarrow$Ar & Pr$\rightarrow$Cl & Pr$\rightarrow$Rw & Rw$\rightarrow$Ar & Rw$\rightarrow$Cl & Rw$\rightarrow$Pr & Avg \\ \midrule
SO & 34.9 & 50.0 & 58.0 & 37.4 & 41.9 & 46.2 & 38.5 & 31.2 & 60.4 & 53.9 & 41.2 & 59.9 & 46.1 \\
DANN~\cite{ganin2016domain} & 45.6 & 59.3 & 70.1 & 47.0 & 58.5 & 60.9 & 46.1 & 43.7 & 68.5 & 63.2 & 51.8 & 76.8 & 57.6 \\
CDAN~\cite{long2018conditional} & 50.7 & 70.6 & 76.0 & 57.6 & 70.0 & 70.0 & 57.4 & 50.9 & 77.3 & 70.9 & 56.7 & 81.6 & 65.8 \\
MMDIA~\cite{jiang2020implicit} & 56.2 & \underline{77.9} & \underline{79.2} & \underline{64.4} & \underline{73.1} & \underline{74.4} & \underline{64.2} & 54.2 & \underline{79.9} & 71.2 & \underline{58.1} & \underline{83.1} & \underline{69.5} \\
MME~\cite{saito2019semi} & 54.2 & 72.8 & 78.3 & 57.9 & 70.2 & 71.8 & 58.5 & 52.9 & 77.9 & \underline{72.7} & 58.1 & 81.8 & 67.3 \\
CDS / MME~\cite{kim2020cross} & \underline{56.9} & 73.3 & 76.5 & 62.8 & 73.1 & 71.1 & 63.0 & \underline{57.9} & 79.4 & 72.5 & 62.5 & 83.0 & 69.3 \\
\Gd PCS (Ours) & 55.8 & 76.9 & \textbf{80.3} & \textbf{67.9} & \textbf{74.0} & \textbf{75.7} & \textbf{67.0} & 52.9 & \textbf{81.0} & \textbf{74.5} & \textbf{58.3} & 82.8& \textbf{70.6} \\ \bottomrule
\end{tabular}%
}
\label{tab:officehome_full}
\end{table*}

\begin{table}[t]
\centering
\caption{Adaptation accuracy (\%) comparison on fully-labeled setting on the Office dataset.}
\resizebox{0.97\linewidth}{!}{%
\begin{tabular}{l|ccccccc}
\toprule
Method & A$\rightarrow$D & A$\rightarrow$W & D$\rightarrow$A & D$\rightarrow$W & W$\rightarrow$A & W$\rightarrow$D & Avg \\ \midrule
SO & 68.9 & 68.4 & 62.5 & 96.7 & 60.7 & 99.3 & 76.1 \\
DANN~\cite{ganin2016domain} & 79.7 & 82 & 68.2 & 96.9 & 67.4 & 99.1 & 82.2 \\
CDAN~\cite{long2018conditional} & \underline{92.9} & \underline{94.1} & 71 & 98.6 & 69.3 & \underline{100} & 87.7 \\
MMDIA~\cite{jiang2020implicit} & 92.1 & 90.3 & 75.3 & \underline{98.7} & \underline{74.9} & 99.8 & \underline{88.8} \\
MME~\cite{saito2019semi} & 88.8 & 87.3 & 69.2 & \underline{98.7} & 65.6 & \underline{100} & 84.9 \\
CDS + MME~\cite{kim2020cross} & 86.9 & 88.3 & \underline{75.9} & 98.6 & 73.3 & \underline{100} & 87.1 \\
\Gd PCS (Ours) & \textbf{94.6} & 92.1 & \textbf{77.4} & 97.7 & \textbf{77.0} & 99.8 & \textbf{89.8} \\ \bottomrule
\end{tabular}%
}
\label{tab:office_full}
\end{table}

\section{Image Retrieval Results}

We present cross-domain image retrieval results in Figure~\ref{fig:retrieval}. Given a query feature $\mathbf{f}_q$ in the target domain, we measure the pairwise cosine similarity between $\mathbf{f}_q$ and all features in the source domain. The source images with the most similar features as $\mathbf{f}_q$ are returned as the top retrieval results. We compare image retrieval results of PCS with CDS in Figure~\ref{fig:retrieval}. As shown in Figure~\ref{fig:retrieval}, features from model trained with CDS are biased to some wrong attributes, \textit{e.g.} color, texture and other visual clues; and quantitatively similar features do not correspond to semantically similar images in different domains. In contrast, we can see that PCS could extract features that are more discriminative and semantically meaningful across domains.

\section{More Ablation Study Results}
In this section, we provide more ablation study results. Ablation experiments similar to Table 2 in the main paper are performed on Office-Home, with results shown in Table~\ref{tab:officehome_ablation}. As we can see in the table, adding each component contributes to the final adaptation accuracy without any performance degradation, which demonstrates the effectiveness of all components in our PCS framework. 

\section{Performance Comparison with UDA Methods using Full Source Labels}

We have shown the superiority of PCS in label-scarce setting (FUDA), and we further conduct experiments with fully-labeled source domain (UDA). The performance comparison with other UDA methods on Office and Office-Home are presented in Table~\ref{tab:office_full} and Table~\ref{tab:officehome_full}, respectively. We can see that PCS achieves the best results even with fully-labeled source, which demonstrates that the proposed PCS could potentially be applied to a wider range of domain adaptation settings.